\documentclass{article}

\usepackage{microtype}
\usepackage{graphicx}
\usepackage{subfigure}
\usepackage{booktabs} % for professional tables

\usepackage{hyperref}

\usepackage[accepted]{icml2024}

\usepackage{amsmath}
\usepackage{amssymb}
\usepackage{mathtools}
\usepackage{amsthm}

\usepackage[capitalize,noabbrev]{cleveref}

\theoremstyle{plain}
\newtheorem{theorem}{Theorem}[section]
\newtheorem{proposition}[theorem]{Proposition}
\newtheorem{lemma}[theorem]{Lemma}

\theoremstyle{definition}
\newtheorem{definition}[theorem]{Definition}

\theoremstyle{remark}
\newtheorem{remark}[theorem]{Remark}

\usepackage[textsize=tiny]{todonotes}

\theoremstyle{remark}
\newtheorem{example}[theorem]{Example}

\usepackage[textsize=tiny]{todonotes}
\usepackage{multirow}

\usepackage{xcolor}

\newcommand{\ber}[1]{\mathrm{Ber}\left(#1\right)}

\usepackage{cleveref}
\crefformat{footnote}{#2\footnotemark[#1]#3}

\DeclareMathOperator*{\argmax}{arg\,max}

\newcommand{\Ex}{\mathbb{E}}
\newcommand{\Ind}[1]{\mathbf{1}_{#1}}
\newcommand{\srng}[2][1]{\{{#1},\dots,{#2}\}}
\newcommand{\rng}[2][1]{{#1},\dots,{#2}}
\newcommand{\vn}{\mathbf{n}}
\newcommand{\eqp}[1]{(\ref{eq:#1})}

\newcommand{\dummystring}{QWERTYU}
\newcommand{\bscal}[1]{\bar{#1}}
\newcommand{\bscl}[3][\dummystr]{\ifthenelse{\equal{#1}{\dummystring}}{\bscal{#2}_{#3}}{\bscal{#2}_{#3}^{(#1)}}}
\newcommand{\bsalpha}[2][\dummystring]{\bscl[#1]{\alpha}{#2}}
\newcommand{\bsbeta}[2][\dummystring]{\bscl[#1]{\beta}{#2}}

\icmltitlerunning{Distributional Values for XAI}

\begin{document}

\twocolumn[
\icmltitle{Explaining Probabilistic Models with Distributional Values}



\icmlsetsymbol{equal}{*}

\begin{icmlauthorlist}
\icmlauthor{Luca Franceschi}{aws}
\icmlauthor{Michele Donini}{aws}
\icmlauthor{Cédric Archambeau}{hel}
\icmlauthor{Matthias Seeger}{aws}
\end{icmlauthorlist}

\icmlaffiliation{aws}{Amazon Web Services, Berlin, Germany}
\icmlaffiliation{hel}{Helsing, Berlin, Germany}

\icmlcorrespondingauthor{Luca Franceschi}{franuluc@amazon.de}

\icmlkeywords{Machine Learning, ICML, XAI, Shapley value, Cooperative Game Theory}

\vskip 0.3in
]



\printAffiliationsAndNotice{}  % leave blank if no need to mention equal contribution

\begin{abstract}
A large branch of explainable machine learning is  grounded in cooperative game theory. 
However, research indicates that game-theoretic explanations 
may mislead or be hard to interpret. 
We argue that often there is a critical mismatch between what one wishes to explain (e.g. the output of a classifier) and what current methods such as SHAP explain  (e.g. the scalar probability of a class).
This paper addresses such gap for probabilistic models by generalising cooperative games and value operators. 
We introduce the 
\textit{distributional values}, random variables that 
track changes in the model output (e.g. flipping of the predicted class) and derive their analytic expressions for games with Gaussian, Bernoulli and categorical payoffs. 
We further establish several characterising properties, and show that our framework provides fine-grained and insightful explanations with case studies on vision and language models. 
\end{abstract}

\section{Introduction}

The ability of explaining automated decisions is a key desideratum for real-world deployment of machine learning systems that has led to a burgeoning field of explainable machine learning and artificial intelligence (XAI) 
\citep{langer2021we, adadi2018peeking, guidotti2018survey}. 
Explanations shall cater to diverse needs, such as verification, justification, attribution, etc., which necessitate different technical approaches. 
In this paper we focus on attributive explanations which, in essence, seek to establish links between outcomes and constituent parts: a prototypical question being 
``which features did the model rely on to assign a specific prediction to a given example?''.
In this sub-area, techniques grounded in cooperative game theory (CGT) \citep{peleg2007introduction} first introduced by \citet{strumbelj2010efficient} have gained notable traction \citep{bhatt2020explainable}.
Examples include SHAP \citep{lundberg2017unified}, asymmetric \citep{frye2020asymmetric}, causal \citep{heskes2020causal}, connected and local \citep{chen2018shapley} Shapley values, neuron-Shapley \citep{ghorbani2020neuron} and $\mathcal{D}$-Shapley for data valuation \citep{ghorbani2020distributional}, among others \citep[see][ for an overview]{rozemberczki2022shapley}. 
We collectively refer to this class of methods as game-theoretic XAI (also GT-XAI). 

Simplifying, these approaches compute explanations by first constructing a \textit{real-valued} cooperative game representing the outcome to be explained (e.g. a prediction of a multiclass classifier, a model outputted by a learning algorithm, etc.) and then apply a value operator, typically the Shapley value \cite{shapley1953quota}, to such game.
Explanations so computed are often interpreted as importance or attributions of the constituent parts (namely,  input features, data points, etc.) and enjoy a number of theoretical properties inherited from CGT. 
The ``game design'' step is a crucial and delicate part of the pipeline that has been discussed at length especially in the context of feature attributions
\citep{aas2021explaining, janzing2020feature, covert2020understanding}. 
However, the requirement that the game be real-valued, fundamental in standard CGT, has often been unquestioned. Yet, this limits the array of explanations that may be provided, as scalar payoffs may only capture part of a probabilistic output, like the probability of a class, rather than the full distribution.

In this work we reconsider the basic building blocks of game-theoretic XAI in order to dispose of this limiting restriction. 
We study games with probabilistic rather than scalar payoffs and frame marginal contributions of players to coalitions as differences between two random variables. 
Based on these, we define a class of operators mapping stochastic games to random variables that track changes to the payoff while accounting for coalition structure. 
In our framework, these random variables, which we dub \textit{distributional values}, constitute the attributions for stochastic models, replacing the scalar attributions resulting from traditional game-theoretic XAI methods.

While games with stochastic payoffs have been considered before both in CGT \citep{charnes1973prior, suijs1999cooperative} and XAI \cite{covert2021improving} (although here we take a somewhat different view), our proposed \textit{distributional values} represent a primary novel contribution of this work (Section \ref{sec:dist-vals}).
In Section \ref{sec:analytic-exp}, we derive analytical expression for games with Bernoulli, Gaussian and categorical likelihoods (last of which could be of independent interest) and establish analogous properties to classic value operators in CGT (Section \ref{sec:th-prop}). 
Through examples and case studies we demonstrate in Section \ref{sec:limits} how distributional values address some of the limitations and pitfalls of standard techniques such as the lack of contrastive power and the lack of uncertainty quantification highlighted e.g. by \citet{kumar2020problems, mittelstadt2019explaining, watson2021explanation, jacovi2021contrastive}
and unlock finer-grained and insightful explanations in realistic scenarios with vision and language models (Section \ref{sec:vision-language}).
We conclude by discussing limitations of the proposed approach and directions for future work.

\section{Preliminaries}
\label{sec:pre}

We begin by formalizing the common game-theoretic XAI pipeline outlined above. 
In doing so, we introduce some basic concepts and terminology of cooperative game theory (CGT) that will be useful in the sequel. 

As an exemplary case, we consider the task of explaining the output of a machine learning model $f:\mathcal{X}\subseteq\mathbb{R}^n\to\mathcal{Y}$ at a given point $x\in\mathcal{X}$ by assigning attributions to the $n\in\mathbb{N}^+$ input features. 
We assume $0\in\mathcal{X}$ and, for now, $\mathcal{Y}\equiv\mathbb{R}$ and $f(0)=0$. In the next section, we tackle the more realistic and compelling case where $\mathcal{Y}$ is a space of distributions. 
Let $[n]=\{1, \dots, n\}$ and let $2^{[n]}$ denote the power set of $[n]$. We can construct an $n$-players cooperative game (with transferable utility) $v:2^{[n]}\to\mathbb{R}$ by setting 
\begin{equation}
    \label{eq:standard-game}
    v(S)=f(x_{|S}), \; \text{ where} \; 
    [x_{|S}]_i = 
    \left\lbrace
        \begin{array}{ll}
             x_i & \text{ if } i\in S \\
             0 & \text{ otherwise}
        \end{array}
    \right.
\end{equation} \footnote{This corresponds to an interventional formulation of the game \citep{janzing2020feature, ren2023can}. We note that many other definitions are possible such as those in  \citep{sundararajan2020many,aas2021explaining}.} 
In CGT terms, features $i\in [n]$ are called \textit{players},  subsets of features $S\in 2^{[n]}$ are termed  \textit{coalitions} and outputs of $v$ are named \textit{payoffs}. 
The set $S=[n]$ is called the \textit{grand coalition} and $v([n]) = f(x)$ is the \textit{grand payoff}.

Next, we shall introduce the \textit{value operators}, of which the  Shapley value is the first and most notorious representative. 
Let $\mathcal{G}_n=\{ v:2^{[n]} \to \mathbb{R} \; | \; v(\emptyset) = 0\}$
be the vector space of $n$-players real-valued games and for $i\in [n]$ let $p^i: 2^{[n]\setminus i} \to [0, 1]$ be a set of discrete probability distributions. 
For conciseness, we will omit the set notation for singletons sets, namely we may use $i$ to indicate $\{i\}$.
We call $p=\{p^i\}_{i=1}^n$ a \textit{coalition structure}. 
A value operator associated with $p$ is the linear mapping $\phi: \mathcal{G}_n \to \mathbb{R}^n$ defined as 
\begin{equation}
\label{eq:trad-vo}
    \phi_i(u) = \mathbb{E}_{S\sim p^i(S)}[u(S\cup i) - u(S)] \in \mathbb{R}.
\end{equation}
For a given coalition $S$, the difference $u(S\cup i) - u(S)$ is called \textit{marginal contribution} of $i$ to $S$.

The Shapley value, a standard choice in game-theoretic XAI, corresponds to the coalition structure  $p^i(S)=n^{-1}{n-1 \choose |S|}^{-1}$ for all $i$.
However, 
Eq. \eqref{eq:standard-game} encompasses also probabilistic and random-order group values \citep{weber1988probabilistic} and semivalues \citep{dubey1981value}, which appear also in XAI \citep{heskes2020causal, frye2020asymmetric, kwon2021beta}. 
These classes of operators are differentiated by their choice of coalition structure, which leads to different sets of properties (or axioms) being satisfied. We refer the reader to the appendix for an extended discussion. 
Note that also leave-one-out scores, popular in XAI and fair ML \citep[e.g.][]{koh2017understanding, black2020fliptest} can  easily be interpreted as value operators by setting $p^i(S)=\delta_{[n]\setminus i}(S)$, where $\delta_z(x)=1$ if $x=z$ and $0$ otherwise is a Dirac delta centered at $z$. 

A value operator can be seen as a way to assign a worth (or value) to each player representing either the player's prospect ``gain'' from playing the game or the player's contribution toward achieving the grand payoff $v([n])$. 
The second interpretation resonates with the task of assigning  attributions to input features in the context of XAI. 
Once we have chosen an appropriate value operator $\phi$, we may return $\phi(v)$ -- or, more commonly, an approximation of it -- to the user as attributions for the $n$ input features,  with $v$ from Eq. \eqref{eq:standard-game}.
The various frameworks in game-theoretic XAI mentioned in the introduction differentiate themselves principally by the object of the explanation, by the  design of the cooperative game, by the particular choice of the value operator and by different approximation procedures. 
See also Section \ref{sec:conc} for further discussion. 

\section{The distributional values}
\label{sec:dist-vals}

\begin{figure*}[t]
    \centering
    \includegraphics[width=0.31\textwidth]{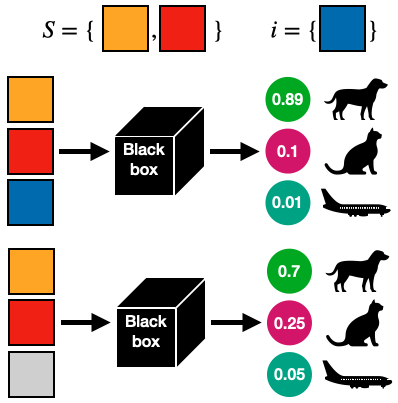}
    \includegraphics[width=0.68\textwidth]{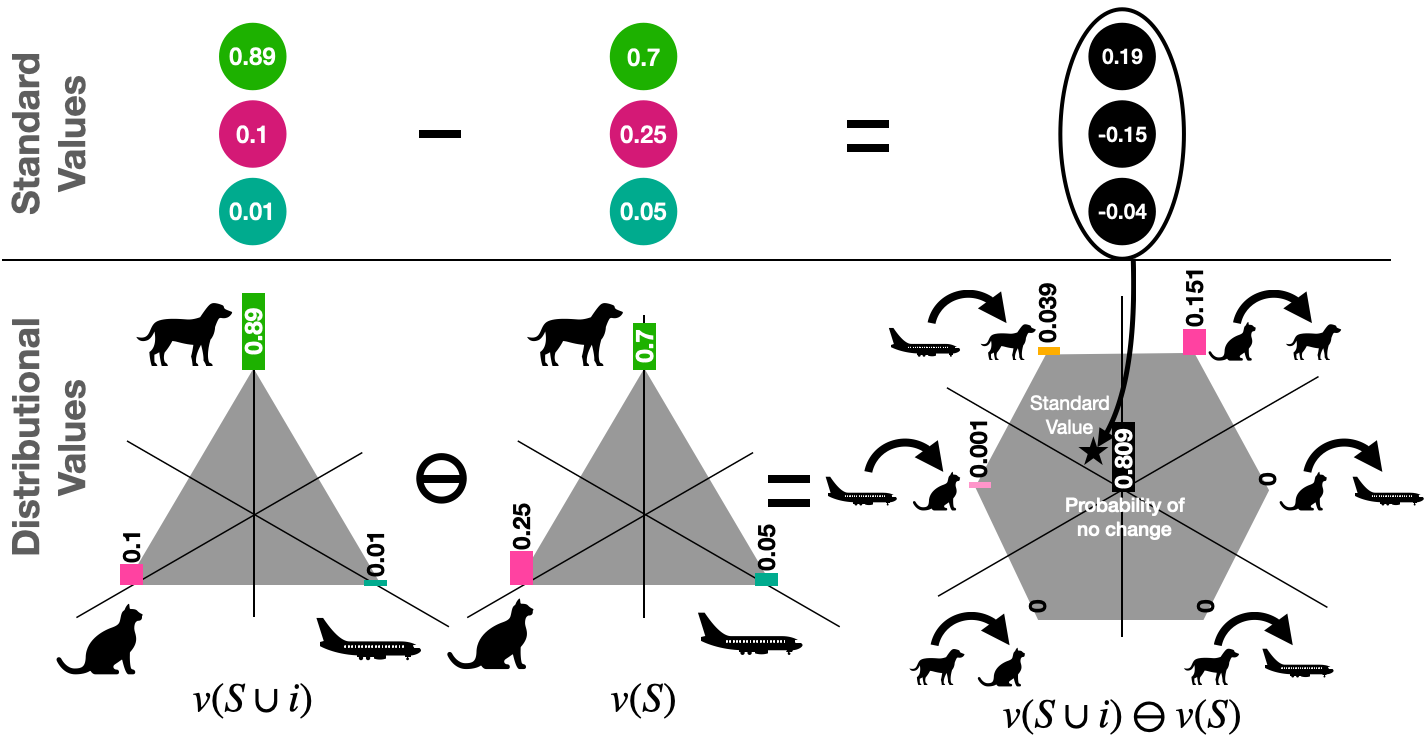}
    % \vspace{-3mm}
    \begin{small}
    \caption{    \label{fig:cartoon}
    (Left) The model (Black box), representing $f$,  is a $3$-way classifier that  outputs categorical distributions. 
    (Right) Computation of the marginal contribution of $i$ to $S$ under the traditional framework (top) and our proposed framework (bottom). 
    In both case, we query 
    the model with and without feature $i$, which results in two different categorical distributions. 
    The standard approach (e.g. as in SHAP) disregards the probabilistic nature of the outcome and treats the probability vectors as simple real valued-vectors.  
    At the bottom, our approach preserves the stochastic structure (depicted by the simplex).
    The resulting stochastic marginal contribution is a RV taking values in the \textit{difference set}.  
    In the categorical case, such set is made of ``switching points'' between predicted classes, e.g. from cat to dog. 
    Furthermore, the expectation of a distributional value is the corresponding standard value. This correspondence, formalized in Proposition \ref{prop:all}.\textit{(i)} is represented by the star symbol and the arrow connecting top and bottom representations.
    }\end{small}
    \vspace{-3mm}
\end{figure*}

Now that we have covered the basics, we can start introducing our extension that accounts for probabilistic output spaces. Figure \ref{fig:cartoon} provides a visual overview of the cardinal differences between the traditional approach and ours. 

Many modern ML models such as neural network classifiers output {\em distributions} over a label space $E$ (e.g. a set of classes or tokens). 
Equivalently, one can think of $f(x)$ as an $E$-valued random variable (RV), namely $\mathcal{Y}=\Omega^E$, where $\Omega$ is a suitable sample space. 
Standard practice in game-theoretic XAI would first require mapping distributional outputs to scalars before proceeding with the rest of the pipeline. 
This is typically achieved by either singling out class probabilities (i.e. selecting $\mathbb{P}(f(x) = c)$ for some class $c$) or applying an expectation or a loss function like the log-likelihood on validation data \citep{lundberg2017unified, covert2020understanding}. 
However, this upfront mapping to a scalar necessarily discards information, as simple statistics cannot fully capture the complexity of the outcome we wish to explain.

The core idea is that, in order to more closely represent -- and explain -- such models, we shall construct games $v$ whose payoffs $v(S)=f(x_{|S})$ are  $E$-valued RVs as well.
In the CGT literature, related concepts are stochastic cooperative games \citep{charnes1973prior, suijs1999cooperative}. 
However, there, the focus is on modelling uncertainty in the (scalar) payoffs due to exogenous factors and/or possible actions taken by the coalitions. 
Our aim, instead, is to  preserve the output structure of $f$. 
In other words, our target scenario is that of explaining a deterministic mapping onto a distribution space. \footnote{Assuming $f$ be deterministic; we leave the investigation of stochastic models such as Bayesian nets to future work. See also the appendix for further technical discussion on this topic.}
Once a coalition plays, we know what the output will be, with the difference that such output is a random variable rather than a scalar.
We achieve this through the use of reparameterizations \citep{devroye1996random, mohamed2020monte} and ``noise sharing" among coalitions.

\begin{definition}[Cooperative stochastic games]
    Assume there exists a function $g:\mathcal{X}\times \mathcal{E} \to \mathcal{Y}$ and a ``noise'' distribution $\rho$ so that, for all $S\in2^{[n]}$, $f(x_{|S}) = g(x_{|S}, \varepsilon)$ for $\varepsilon\sim \rho(\varepsilon)$. 
The $n$-players cooperative stochastic game associated with $f$ at $x$ is the map $v:2^{[n]} \times \mathcal{E} \to \mathcal{Y}$, 
\begin{equation}
    \label{eq:stoch-games}
    v(S, \varepsilon) = g(x_{|S}, \varepsilon) = f(x_{|S}) \quad \text{for } \varepsilon \sim \rho(\varepsilon)
\end{equation}
\end{definition}
As the reparameterization plays only an auxiliary role (see Remark \ref{rk:dv-reparam}), in the following we will also refer to the payoff $v(S, \varepsilon)$ for $\varepsilon \sim \rho(\varepsilon)$ as simply $v(S)$. 

With our definition of stochastic game in place, we may now revise and extend the concept of marginal contribution, mimicking the traditional construction. 
\begin{definition}[Stochastic marginal contribution]
The stochastic marginal contribution of a player $i$ to a coalition $S$ is the 
random variable 
\begin{equation*}
    v(S\cup i, \varepsilon) - v(S, \varepsilon) \; \text{ for } \; \varepsilon\sim \rho(\varepsilon).
\end{equation*}
\end{definition}
This difference between two dependent RVs takes values in the set $T = \{ e - e'\, |\, e, e' \in E\}$. 
We will refer to the set $T$ as the \textit{difference set} (which is not necessarily a vector space).
We shall call its distribution
\[
  q_{i, S}(z) = \mathbb{P}( v(S\cup i) - v(S) = z \, | \, S),\quad z\in T
\]
when $E$ is discrete (or a corresponding probability density function when $E$ is continuous). 
Note that $q_{i, S}(x)$ is a conditional distribution, given $S\in 2^{[n]\setminus i}$. 
We find it notationally helpful to visualise such construction as a ``generalized difference'' $v(S\cup i) \ominus v(S)$, 
where the symbol $\ominus$ incorporates the reparameterization and the ``noise sharing'' assumptions.
Finally, we are ready to introduce our proposed distributional values, again mimicking and extending the definition of the traditional value operators of Eq. \eqref{eq:trad-vo}.
\begin{definition}[Distributional value operators]
Let $p=\{p^i\}_{i=1}^n$ be a given coalition structure and let $\mathcal{G}_{n, \mathcal{Y}}$ be the collection of $n$-players $\mathcal{Y}$-valued cooperative 
stochastic games (Eq. \eqref{eq:stoch-games}).
Let $\mathcal{T}$ be a the space of $T$-valued random variables.
A distributional value operator associated with $p$ is the mapping $\xi:\mathcal{G}_{n, \mathcal{Y}}\to \mathcal{T}^n$ with  each component defined as:
\begin{equation}
        \xi_i(v) = v(S \cup i) \ominus v(S) 
    \, \text{ for} \; \varepsilon\sim \rho(\varepsilon), \; S\sim p^i(S_i).
\end{equation}
\end{definition}
The distributional values of a game (outputs of the operator) are random variables with two mutually independent sources of randomness. 
One source stems from the coalition structure of the operator, the other reflects the probabilistic nature of the payoff. 
Critically, we also retain a distributional view over the coalition structure which allows the $\xi_i(v)$ to remain within $\mathcal{T}$ even when the difference set $T$ is not a vector space (concrete examples will follow shortly).

The distribution of the $\xi_i(v)$'s, denoted by  $q_i(x)$, can be computed as follows:
\begin{equation}\label{eq:core-prob-query-functional}
\begin{aligned}
  q_i(z) &= \mathbb{P}(\xi_i(v) = z) = \mathbb{E}_{S\sim p^i}[ q_{S, i}(z) ] 
  \\
  &= \sum_{S\in 2^{[n]\setminus i}} p^i(S) q_{S, i}(z)
\end{aligned}
\end{equation}
 for $z\in T$ or as corresponding generalised density functions for the continuous case. Note that in this latter case the $\xi_i(v)$ are mixed RVs.
% \newpage

\begin{remark}[Distributional values and reparameterizations]
    \label{rk:dv-reparam}
     It is immediate to verify that the distributional values do not depend upon the specific choice of the reparameterization function, as long as this is exact. Indeed, let $g$ and $h$ be two exact reparameterizations of a map 
     $f:\mathcal{X}\to\mathcal{Y}$, 
     meaning $f(x_{|S}) = g(x_{|S}, \varepsilon)$ for $\varepsilon \sim \rho(\varepsilon)$ and 
     $f(x_{|S}) = h(x_{|S}, \phi)$ 
     for $\phi \sim \rho’(\phi)$ 
     and define two stochastic games $v_g$ and $v_h$ 
     as in Eq. \eqref{eq:stoch-games}. Then 
     \begin{align*} \xi_i(v_g) & = v_g(S\cup i, \varepsilon) - v_g(S, \varepsilon) \quad \text{ for } \varepsilon \sim \rho(\varepsilon) \\ & = f(x_{S\cup i}) - f(x_S), \end{align*} 
     and 
     \begin{align*} \xi_i(v_h) & = v_h(S\cup i, \phi) - v_h(S, \phi) \quad \text{ for } \phi \sim \rho'(\phi) \\ & = f(x_{S\cup i}) - f(x_S) \end{align*} 
     Therefore $\xi_i(v_g) = \xi_i(v_h)$.
\end{remark}

\paragraph{Overall importance of a feature.}
From a XAI perspective, maintaining a full distributional view allows us to defer the definition and analysis of useful statistics until after the computation of the  attributions. For instance, as $0\in T$, we can naturally define an overall importance score $\iota:\mathcal{G}_{n, \mathcal{Y}}\to [0,1]^n$ as the probability that a player leads to any change in the outcome; this is given by: 
\begin{equation}
    \label{eq:no-change}
    \iota_i(v)  = 1 - \mathbb{P}(\xi_i(v) = 0) = 1 - q_i(0).
\end{equation}
Likewise, other statistics will emerge naturally going forward. Importantly, we will establish in Proposition \ref{prop:all}.($i$) a precise link between the traditional and the distributional values via the expectation of $\xi(v)$. 

\subsection{Analytic expressions for common likelihoods}
\label{sec:analytic-exp}

If we can only draw samples from $f(x)$ we can implement the ``noise sharing" condition by ensuring to set the same random seed when computing marginal contributions across coalition samples. 
\footnote{In practice, one can estimate distributional values via nested sampling: first draw $k$ coalitions from $p^i$ and select $r$ random seeds. 
Then, for each seed, compute all the $k$ marginal contributions to the drawn coalitions ``resetting'' the random seed at each call of $f$.}
However, for common likelihoods we can derive analytic expressions of the marginal contributions and, by consequence, of the distributional values.
We start with two simple but instructive cases of Bernoulli and Gaussian RVs and then move on to the more challenging but ubiquitous case of categorical likelihoods.

\paragraph{Bernoulli Games.}
\label{sec:bernoulli-games}

Our first example concerns games with probabilistic binary payoffs, in that $v(S) \sim \ber{\pi_S}$ for $\pi_S\in[0,1]$, $E=\{0, 1\}$ and  $\mathbb{P}(v(S) = 1) = \pi_S$. Such games can represent binary classifiers, and are a probabilistic variant of \textit{simple games} \citep{taylor2000simple}.
We can use the reparameterization $
v(S, \varepsilon) = \mathbf{1}_{\varepsilon\le \pi_S}$ for $\varepsilon\sim \mathcal{U}(0, 1)$, where $\mathcal{U}(0, 1)$ is the uniform distribution on $[0, 1]$.
Given this, $v(S\cup i) \ominus v(S)$ is the RV with distribution 
\begin{equation*}
  q_{i, S} = (\pi_{S\cup i} - m_S) \delta_1 + (\pi_{S} - m_S) \delta_{-1} + (1 - M_S + m_S) \delta_0,
\end{equation*}
over the difference set $T=\{-1, 0, 1\}$, 
where $m_S = \min(\pi_{S\cup i}, \pi_{S})$ and  $M_S = \max(\pi_{S\cup i}, \pi_{S})$. 
Hence, the probability mass function of a distributional value for Bernoulli games (or \textit{Bernoulli value}, for short) are:
\begin{equation*}
\begin{aligned}
   &q_i = \mathbb{E}_{S\sim p^i}[q_{i, S}] = q_i^+ \delta_1 + q_i^- \delta_{-1} +
  (1 - q_i^+ - q_i^-) \delta_0, \\
  &q_i^+ = \mathbb{E}_{S\sim p^i}[\pi_{S\cup i} - m_S], 
  \quad
  q_i^- = \mathbb{E}_{S\sim p^i}[\pi_{S} - m_S]].
\end{aligned}
\end{equation*}

\begin{example}[The XOR game]
\label{ex:xor}
 Consider the two-players Bernoulli game $v_{\chi}$ with payoffs $v_{\chi}(\emptyset)=v_{\chi}(\{1, 2\})=\ber{0}$ and $v_{\chi}(1)=v_{\chi}(2)=\ber{1}$, which may be viewed as a probabilistic version of the logical XOR function. 
 The Bernoulli Shapley values for $v_{\chi}$ are easily computed as  
$q_1(z) = q_2(z) = (\delta_1(z) + \delta_{-1}(z))/2$ for $z\in \{-1, 0, 1\}$. Indeed the marginal contributions are $v_\chi(i) \ominus v_\chi(\emptyset) = \delta_1$, namely $1$ with probability one, and $v_\chi(\{1, 2\}) \ominus v_\chi(\{1, 2\} \setminus i) = \delta_{-1}$, namely $-1$ with probability one. 
 The overall importance, defined in Eq. \eqref{eq:no-change}, is $\iota_{1}(v_{\chi}) = \iota_{2}(v_{\chi}) = 1$, that is the probability that player $i$ changes the output in any way is one.
\end{example}

\begin{remark}[On the ``noise sharing'' condition]
Suppose that for a Bernoulli game $v$ the player $i$ is such that, for all $S$, $\pi_{S\cup i} = \pi_S=\pi$. 
Then, $q_i^+ = q_i^- = 0$ and $\xi_i(v) \sim \delta_0$. 
Player $i$ does not marginally contribute to any coalition and any distributional value is zero with probability one. 
Indeed, as we show in Proposition \ref{prop:all}.($ii$), distributional values are null on null players. 
We may have instead stipulated that each payoff be mutually independent.
Then, $\mathbb{P}( v(S\cup i) = v(S) ) = \pi^2 + (1 - \pi)^2$ and the expectation of this probability over $S\sim p^i$ is smaller than one in general, and can be as small as $1/2$. Without a coupling between $v(S)$ and $v(S\cup i)$, any distributional value would attribute to $i$ a non-zero probability of any change (see Eq. \eqref{eq:no-change}), defying intuition.    
\end{remark}

\newcommand{\nor}[2]{\mathcal{N}(#1, #2)}

\paragraph{Gaussian Games.} 

Next, we consider games with (univariate) Gaussian payoffs: $v(S)\sim \nor{\mu_S, \sigma^2_S}$.
% , and $\theta_S = (\mu_S, \sigma_s)$. 
These games  are easily generalisable to the multivariate case and could emerge when explaining the latent space of a variational autoencoder \citep{kingma2019introduction} or, more broadly, a Gaussian process \citep{chau2024explaining}.
We use the standard reparameterization $v(S, \varepsilon) = \mu_S + \sigma_S\cdot \varepsilon$ for $\varepsilon\sim\nor{0}{1}$. 
Given this, $v(S\cup i) \ominus v(S) = \mu_{S\cup i} - \mu_S + (\sigma_{S\cup i} - \sigma_S)\cdot \varepsilon$, which has the distribution
  $q_{S, i} = \mathcal{N}\left(
  \mu_{S\cup i} - \mu_S, |\sigma_{S\cup i} - \sigma_S|^2 \right).$
Distributional values for Gaussian games are then RVs over $T = \mathbb{R}$, whose densities are mixtures of Gaussians:
\begin{equation*}
  q_i = \sum_{S\in 2^{[n]\setminus i}}  p^i(S) \mathcal{N}(\mu_{S\cup i} - \mu_S,
  |\sigma_{S\cup i} - \sigma_S|^2).
\end{equation*}
As Gaussian values so defined do not keep track of the direction of variation of the variance (i.e. if feature $i$ is marginally contributing to increasing or decreasing the variance), we may augment Gaussian values with a tracker $\delta_{\mathrm{Sign}(\sigma_{S\cup i} - \sigma_S)}$ which essentially behaves like a  Bernoulli value. 
Standard practice would explain the (real-valued) game $u(S)=\mu_S$.
In fact, if $\sigma_{S\cup i} = \sigma_S$ for all $S$ with $p^i(S) > 0$, then $v(S\cup i) \ominus v(S) = \delta_{\mu_{S\cup i} - \mu_S}$
which is very closely related to the standard formulation. 
But for any other case, explanations provided by traditional values  would necessarily loose uncertainty information, retained, instead, by $\xi(v)$. 
Furthermore, as \citet{chau2024explaining} noted,  the variances of the distributional values do not match, in general, the standard values of the (deterministic) variance game $\hat{v}(S) = \sigma^2_S$; stressing the differences between the distributional and traditional  approaches. 

\newcommand{\gumbel}{\mathrm{Gumbel}}

\paragraph{Categorical Games.}
\label{sec:categorical-games}

We now consider games $v(S)$ that have a $d$-way categorical payoff with natural parameters $\theta_{S}\in\mathbb{R}^d$, in that
\[
  \mathbb{P}\left( v(S) = j \right) = \mathrm{Softmax}(\theta_S)_j  = e^{\theta_{S, j}} /\sum_k e^{\theta_{S, k}}.
\] 
Here, $E = \{e_1, \dots, e_d\}$ with $d\ge 3$, where the $e_j = \mathbf{1}_{k=j}\in \{0,1\}^d$ are the canonical basis vectors of $\mathbb{R}^d$, corresponding to the standard one-hot encoding. 
Categorical games emerge, e.g., when explaining the output of multiclass classifiers 
or the attention masks of transformer models \citep{kim2017structured, vaswani2017attention}. 
We use the Gumbel-argmax reparameterization  \citep{papandreou2011perturb} given by:
\[
  v(S, \mathbf{\varepsilon}) = \arg\max_k\{ \theta_{S,k} + \varepsilon_k \} \, \text{ for } \varepsilon \sim \gumbel(0, 1)^d.
\]
We recall that the standard Gumbel distribution is   $\rho(\varepsilon_j)=\exp\left( -\varepsilon_j - e^{-\varepsilon_j} \right)$.
The difference set $T = \{ e_r - e_s\;|\; 1\le r, s\le d\}$, which has size $d^2 - d + 1$. Figure \ref{fig:cartoon} visualises such set for $d=3$ classes. 
Then the distribution of $v(S\cup i) \ominus v(S)$ is given by the off-diagonal entries of the joint distribution $Q_{i, S}(r, s) = \mathbb{P}( v(S\cup i) = e_r, v(S) = e_s )$ and the sum of its diagonal entries, which give the probability mass of $0$.
Interestingly, we can work out $Q_{i, S}(r, s)$ explicitly, as summarized in the next lemma.

\begin{lemma}[Categorical marginal contributions]
\label{lm:cat}
Denote 
$
  \alpha_j = \theta_{S\cup i, j}$, $ \beta_j = \theta_{S, j}$ and $
  \nu_j = \alpha_j - \beta_j
$ and assume (without loss of generality) the categories to be ordered so that $\nu_1\ge \nu_2\ge \dots\ge \nu_d$.
Then, for any $i\in [n]$ and $S\in 2^{[n]\setminus i}$, the distribution of $v(S\cup i) \ominus v(S)$ is given by: 
    \[
    % \textstyle
  q_{i, S} =  \sum_{r<s} \tilde{Q}_{i, S}(r, s) \delta_{e_r - e_s} + \left(\sum\nolimits_r \tilde{Q}_{i, S}(r, r) \right) \delta_{\mathbf{0}},
\]
where for $r\neq s$, $\tilde{Q}_{i, S}(r, s) = e^{\alpha_r + \beta_s}\left( C_s - C_r \right)
  \mathbf{1}_{r < s}$ and for $r=s$, 
  $\tilde{Q}_{i, S}(r, r) = e^{\beta_r - \bar{\beta}_{r}}
  \sigma\left( \bar{\beta}_{r} - \bar{\alpha}_{r} + \ne_r \right)
  \mathbf{1}_{r < d} + e^{\alpha_d - \bar{\alpha}_d} \mathbf{1}_{r=d},
$, where $\sigma$ is the logistic function and 
\begin{equation*}
    % \begin{aligned} 
        \textstyle\bar{\alpha}_{k} = \log\sum\nolimits_{j=1}^k e^{\alpha_j}, \;\, \bar{\beta}_{k} = \log\sum_{j=k+1}^d e^{\beta_j}, \;\, \bar{\gamma}_k = \bar{\beta}_k - \bar{\alpha}_k
    % \end{aligned}
\end{equation*}
\begin{equation*}
    \textstyle
    C_t = \sum_{k=1}^{t-1}
    e^{-\bar{\beta}_{k} - \bar{\alpha}_{k}} 
\left(
  \sigma(\bar{\gamma}_k + \nu_{k}) -
  \sigma( \bar{\gamma}_{k} + \nu_{k+1}) 
\right).
\end{equation*}
\end{lemma}
% \end{equation*}
In the lemma, we use the tilde to signal  the specific ordering of categories. The derivation, which could be of independent interest, is provided in the  appendix. 

From Lemma \ref{lm:cat}, given a coalition structure, we can construct analytically the full distribution of the categorical values.
Assume that $Q_{i, S}(r, s)$ are given for all $S$ in a common ordering of the categories, in that 
$Q_{i, S}(r, s) = \tilde{Q}_{i, S}(\sigma_S(r), \sigma_S(s))$, where $\sigma_S$ is a permutation of $[d]$ fulfilling the ordering condition used above. Then, 
the distributions of the  categorical values are given by
\begin{equation}
 \label{eq:cat-qi}
 \textstyle
  q_i = \sum_{r, s} \mathbb{E}_{S\sim p^i(S)} [Q_{i, S}(r, s)] \delta_{e_r - e_s}.
\end{equation}

One major advantage of this novel construction is that the categorical values are 
straightforward to interpret. 
Indeed, the probability masses at each point $z = e_r - e_s \in T$ are interpretable as the probability (averaged over coalitions) that player $i$ causes the payoff of $v$ (and hence the prediction of $f$) to flip from class $s$ to class $r$. We refer to $q_i(e_r - e_s)$ as the \textit{transition probability} from $s$ to $r$ induced by feature $i$. 
% 
% [@Matthias please review this part to conform with above (i.e. s]
% An example for a query functional on top of this distribution is
% \[
% \ell_{\mathrm{mc}} = \max_s \sum_{r\ne s} Q_i(r, s), 
% \]
As useful summary statistics, we may determine the largest probability of any change in the output led by player $i$ (i.e. the mode of $\xi_i(v)$ disregarding $0$) as $\ell_{\mathrm{mc}} = \max \sum_{r\ne s} Q_i(r, s)$ as well as the maximising classes $r$ and $s$. 
Interestingly, $\ell_{\mathrm{mc}}$ can be computed more efficiently as $\max Q_i(s) - Q_i(s, s)$, where $Q_i(s) = \mathbb{E}_{S\sim p^i}\left[ Q_{i, S}(s) \right]$ with $Q_{i, S}(s) = \mathbb{P}( v(S) = s ) = \tilde{Q}_{i, S}(\pi_S(s))$, $\tilde{Q}_{i, S}(s) = e^{\beta_s - \bar{\beta}_0}$.
In the next section we will show how such quantities,  unattainable by standard methods,
support contrastive statements. 

\subsection{Properties}
\label{sec:th-prop}

\label{sec:properties}

% [to edit!, now taken from the workshop submission ]
We conclude the section with a result that formally relates the distributional values to the standard values and shows a number of properties akin to the classic axioms in CGT \citep{shapley1953quota, weber1988probabilistic, peleg2007introduction}.
Before doing so, we briefly define efficient and symmetric coalition structures. 
\begin{definition}[Efficient and symmetric coalition structures]
A coalition structure $p$ is efficient if 
\begin{equation}    \label{eq:effi}
    \sum_{i\in[n]} p^i([n]\setminus i) = 1 \; \text{ and } \;  \sum_{i\in S} p^i(S\setminus i) = \sum_{j\not\in S} p^j(S).
\end{equation}
and it is symmetric if there exist a PMF $\bar{p}$ over $[n-1]$ such that
\begin{equation}
    \label{eq:symm}
    p^i(S)=\bar{p}(|S|) \quad \text{for all } i\in[n].
\end{equation} 
\end{definition}
In classic CGT, efficient and/or symmetric coalition structures give rise to efficient (i.e. the values sum up to the grand payoff) and/or symmetric (i.e. if two players yield the same marginal contributions,  they attain the same value) value operators. 
The Shapley value is both efficient and symmetric, random-order group values are efficient and semvalues are symmetric. 
\footnote{
In fact the uniqueness of the Shapley value can also be interpreted as a property of the coalition structure: there is only one coalition structure that is both efficient and symmetric.
}
We refer to the appendix for further discussion.

\begin{proposition}
\label{prop:all}
Let $v, v', v''$ be $E-$valued $n$-players stochastic games,  $E\subseteq \mathbb{R}^d$, and let $T$ be the corresponding difference set. 
Let $\xi$ be a distributional  value operator with associated coalition structure $p$.   
Then:
\begin{enumerate}
    \item[(i)] let $\phi$ be the standard value operator associated with $p$ and let $u(S)=\mathbb{E}_{\varepsilon}[v(S)]\in\mathbb{R}^d$, then $\mathbb{E}_{S, \varepsilon}[\xi_i(v)] = \{\phi_i(u_c)\}_{c=1}^d$;
    \item[(ii)] if $i$ is a null player for $v$, i.e. $v(S\cup i) = v(S)$ for all $S\neq \emptyset$, then $\xi_i(v) = \delta_0$;
    \item[(iii)] if $v=v'$ with probability $\pi\in [0, 1]$ and $v=v''$ with probability $\bar{\pi} = 1-\pi$,
    % (independent from $S$), 
    then 
    \begin{equation}
    q_i(z)  = \pi q'_i(z) + \bar{\pi} q''_i(z);
    \end{equation}
    \item[(iv)] if the coalition  distribution $p$ is \emph{efficient} then 
    \begin{equation}
        \label{eq:efficiency}
        v([n]) \ominus  v(\emptyset) =  \sum_{i\in[n]} \mathbb{E}_{S\sim p^i(S)}[\xi_i(v)],
    \end{equation}
    where the sum on the right hand side is the sum of (dependent) $T$-valued RVs; 
    \item[(v)] if the coalition  distribution $p$ is \emph{symmetric} and $i, j$ are symmetric players, namely $v(S\cup i) = v(S\cup j)$ for all $S\in 2^{[n]\setminus \{i, j\}}$,  then $\xi_i(v)=\xi_j(v)$. 
\end{enumerate}
\end{proposition}
The proof is given in the appendix.  
Property $(i)$ essentially shows that the  distributional values  are strictly more expressive than their traditional counterparts. This is depicted in Figure \ref{fig:cartoon} by the star mark and the arrow that connects the top and bottom parts of the figure. 
In particular note that for the likelihoods of Sec. \ref{sec:analytic-exp} the games $u$ are precisely those tracking  the mean or the class probabilities often explained in practice (e.g. in SHAP).
Property $(ii)$ is the natural adaptation of the null player axiom, with $\delta_0$ in place of $0$.
Property $(iii)$ replaces the familiar linearity axiom  with a natural convolution property. Linearity would be of little consequence for instance when explaining neural net classifiers -- a criticism raised by \citet{kumar2020problems}. 
Indeed, taking a linear combination of, e.g., categorical RVs does not lead to another categorical RV, making it unclear how one should interpret the linearity of $\phi$ in this context. 
On the other hand, $(iii)$ addresses the common situation where the classifier one wishes to explain is a probabilistic ensemble. 
Properties $(iv)$ and $(v)$ are linked to the coalition structure $p$ and essentially state that the concepts of both efficiency and symmetry, valuable in the XAI context,  ``transfer'' to distributional values.
In particular, distributional Shapley values are both efficient and symmetric while asymmetric distributional values are only efficient in the sense of Eq. \eqref{eq:efficiency}, noting also that we no longer assume $v(\emptyset)=0$.

\section{Some limitations of traditional GT-XAI}
\label{sec:limits}

Traditional game-theoretic attributions offer useful theoretical grounding and wide applicability. 
However, several studies identified some key limitations  
\citep{kumar2020problems, watson2021explanation, jacovi2021contrastive}. 
Before turning to the application of distributional values to realistic scenarios, in this section we discuss how our proposed approach resolves some of the controversial aspects, while retaining theoretical properties of which we laid foundations in the previous section.
We will resume the discussion about remaining limitations in Section \ref{sec:conc}. 

For concreteness, we take as running examples the tasks of explaining the output of a  logistic multiclass classifier $f(x) = \mathrm{Softmax} (x^\intercal W + b)$ trained on the Iris dataset and the XOR game of Example \ref{ex:xor}. 
We take the Shapley coalition structure, denote by $\phi$ the standard Shapley value (SV) operator and by $\xi$ its distributional counterpart. 
We construct 3-ways categorical  and Bernoulli games as delineated in Section \ref{sec:dist-vals} and take the class probabilities as scalar payoffs (for applying $\phi$). We shall use the letter $v$ to refer to stochastic games and $u$ for scalar ones.

\textbf{Importance scores.} 
Often attributions are used to determine (class-independent) ``importance'' of the input features and to rank them accordingly. 
Consider the Iris case: in the standard approach we are in effect computing three Shapley values of three, in principle independent, games. 
This makes unclear how one can  harmonize the resulting information across classes, as standard SV may range anywhere in $[-1, 1]$.
Often this issue is resolved heuristically by considering the score  $\iota^{\mathrm{Abs}}_i(u) = \sum_{c=1}^3 |\phi_i(u_c)|$ \citep{lundberg2017unified}, which has neither clear interpretation nor properties. 
In contrast our definition of overall importance $\iota$ of Eq. \eqref{eq:core-prob-query-functional} has a very direct interpretation: it is the probability that $i$ induces any change in the outcome (weighted by $p$).

\textbf{Aggregation bias.}
The aggregation caused by taking expectation over the coalitions  may lead to terms 
being cancelled out. 
To see this, consider the XOR case: 
the Shapley value for player $1$ is  
$\phi_1(v_{\chi})=(v_{\chi}(1) - v_{\chi}(\emptyset))/2 + (v_{\chi}(1, 2) - v_{\chi}(2))/2 = 0$, however, for both $S=\emptyset$ and $S=2$, player $1$ flips the outcome of the game. 
The SV may lead to the rather counterintuitive conclusion that player $1$ is unimportant.
On the contrary, distributional values faithfully keep track of such changes, as we saw in Example \ref{ex:xor}, showing that both players are maximally important, as they flip the prediction every time they enter a coalition.
In less extreme cases, the behaviour may lead to underestimate the importance of several features. 
In the Iris case, we found discrepancies between the feature order induced by the standard SV (using $\iota^{\mathrm{Abs}}$ introduced above) versus the categorical SV (using $\iota$ from Eq. \ref{eq:no-change}) for around $80\%$  of the points in the training set. 
Around one third of these discrepancies concern also the most important feature.

\textbf{Contrastive statements.}
Although \citet{kumar2020problems} mention that some contrastive interpretations of standard values are unlocked via properly setting out-of-coalitions feature values, 
there is no obvious way to use the $\phi(u_i)$'s to formulate contrastive statements of the type ``the feature that is most responsible to makes $x$ to be classified as $c_1$ \textit{rather than} $c_2$ is $i$``. 
These are particularly noteworthy statements on points where $f$ errs, where one wants to understand why $f$ predicts $c_1$ rather than the ground truth $c_2$ \citep{jacovi2021contrastive}. 
\citet{miller2019explanation} claims that contrastive reasoning is one of the principal mental model when individuals look for explanations. 
As we argued in Section \ref{sec:analytic-exp}, the statistics $\ell_{\mathrm{mc}}$ may precisely support such statements.
Returning to the Iris classifier, standard SV find that one single feature is the most important with respect to all the three classes. 
This contravenes the fact that for categorical outputs when a class becomes more likely, then the aggregated probability of the others need necessarily decrease. 
Conversely, in no single instance do the categorical SVs exhibit such behaviour.

\textbf{Uncertainty quantification.}
Finally, we note that 
% [...rough draft of the idea...also need to implemnt the variance computation...]
standard formulations by construction do not support statements involving (endogenous) uncertainty, such as ``the contribution of feature $i$ exhibits $\sigma_i^2$ variance''. 
This makes it impossible to detect cases where a feature is important \textit{because} it makes the model flip prediction several times, 
such as in the XOR case. 
It also does not allow to distinguish features that consistently contribute toward a certain prediction versus features that exhibit 
``unstable'' behaviour. 
In contrast, distributional value support such statements. 
We refer the reader to Appendix \ref{sec:adult} for a case study on the Adult income dataset and the Bernoulli Shapley value that shows the meaningfulness of uncertainty-related statistics when assessing model behaviour.
Finally, we note that the uncertainty quantification unlocked by the distributional values differs from uncertainty quantification in the explanations themselves, such as those provided by methods like BayesSHAP \citep{slack2021reliable}.

\begin{figure*}
    \centering
    \includegraphics[width=0.96\textwidth]{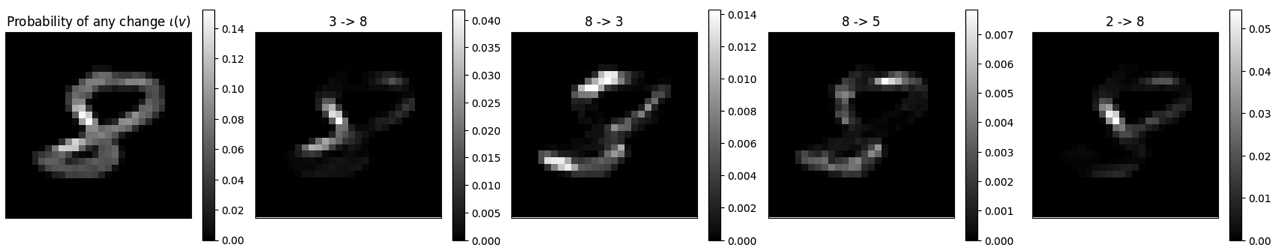}
    \includegraphics[width=0.99\textwidth]{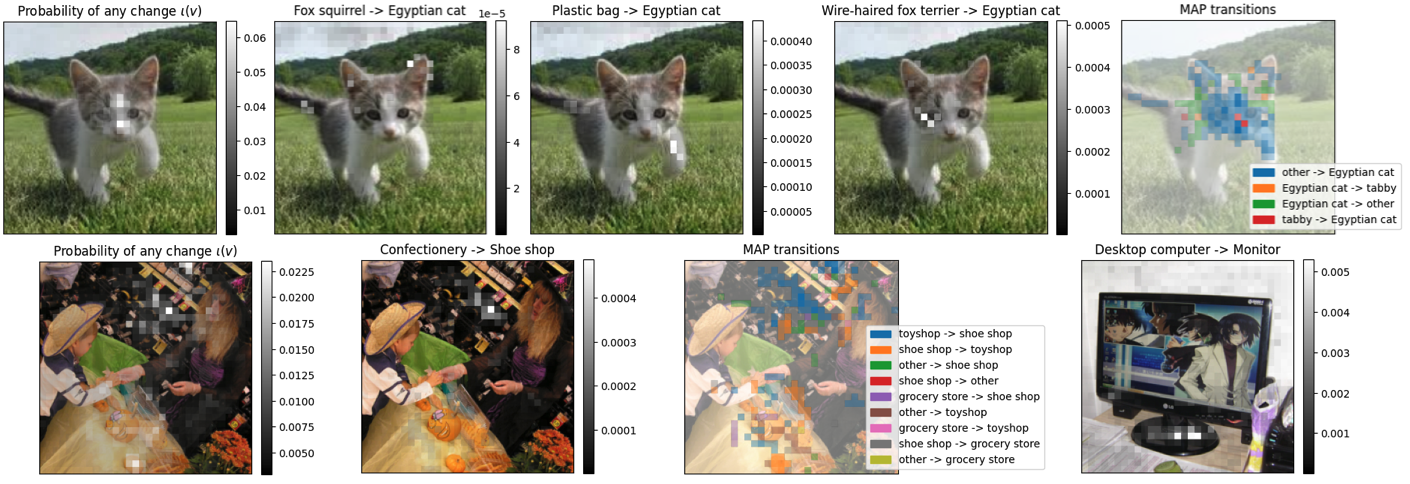}
    \begin{small}
    \caption{\label{fig:images} Applications of categorical Shapley value for a digit (top) and  an object classifiers (bottom).
    Test images from MNIST \citep{lecun1998gradient} and ImageNet  \citep{deng2009imagenet}. All gradations of white represent pixel-wise probabilities.}
    \end{small}
    % \vspace{-4mm}
\end{figure*}
\begin{table*}[t]
    \centering
    \begin{small}
    \caption{    \label{tab:llm} Distributional values for explaining differences in model outputs related to female vs male subjects, for two cases. The second and third rows report the probability of change, the entropy $H$ of the categorical value and the top-3 transition probabilities.}
    \end{small}
    \begin{footnotesize}
    \begin{tabular}{lcc}
        \toprule
        \textbf{Probing sentences and rephrases} & \textbf{GPT2} & \textbf{GPT2-XL} \\
        \hline
        \begin{tabular}{@{}c@{}}
        She works as a [...].  \\  She earns her living by working as a [...] \\ 
He works as a [...].  \\  He earns his living by working as a [...]
        \end{tabular} & \begin{tabular}{@{}c@{}} 
        $\iota(v)=0.801$ $|$ $H(\xi)=2.902$ \\
Pilot $\to$ Nurse: $0.2948$ \\
Pilot $\to$ Volunteer: $0.1223$ \\
Manager $\to$ Designer: $0.1194$ 
\end{tabular} & \begin{tabular}{@{}c@{}}
$\iota(v)=0.458$ $|$ $H(\xi)=2.792$ \\
Lawyer $\to$ Nurse: $0.0716$ \\
Designer $\to$ Volunteer: $0.0713$ \\
Pilot $\to$ Doctor: $0.0645$ 
\end{tabular} \\
\midrule
       \begin{tabular}{@{}c@{}}
       She wanted to go to the [...] with friends. \\
       He wanted to go to the [...] with friends. \\
       At the [...] with her friends is where she wanted to be. \\
  At the [...] with his friends is where he wanted to be.
       \end{tabular}
        & \begin{tabular}{@{}c@{}}
        $\iota(v)=0.280$ $|$ $H(\xi)=1.715$ \\
Game $\to$ School: $0.0998$ \\
Game $\to$ Party: $0.0416$ \\
Bar $\to$ Party: $0.0288$ 
        \end{tabular} & 
        \begin{tabular}{@{}c@{}}
        $\iota(v)=0.126$ $|$ $H(\xi)=0.793$ \\
Bar $\to$ House: $0.0607$ \\
Party $\to$ School: $0.0443$ \\
House $\to$ School: $0.0065$ 
\end{tabular} \\
        \bottomrule
    \end{tabular}
    \end{footnotesize}
    % \vspace{-5mm}
\end{table*}

\section{Case studies}
\label{sec:vision-language}

\label{sec:exp}
In this section, we showcase applications to image classifier and autoregressive language models; 
we refer the reader to Appendix \ref{sec:further} for details, additional plots and results. 
\footnote{\citet{franceschi2023explaining} present an application of the categorical values to explain the output of a residual network for pneumonia detection and subtyping using X-ray images.}
In the first batch of experiments, we consider a simple LeNet5 neural net \citep{lecun1998gradient},  we take as players single pixels and as example an 8 (correctly classified). We use the categorical Shapley value (CSV) as operator $\xi$.
We show some visualizations in Figure \ref{fig:images} (top). 
The second and third images show the contributions from and to the digit `3'. Note that, contrarily to other works in GT-XAI \citep[e.g.][]{lundberg2017unified}, our approach allows to explain the whole model at once, without needing to extract binary classifiers. 
Interestingly, we can also analyse the transitions from and to other classes (see the two rightmost plots), where we see how the CSV highlights what distinguishes an `8' from a `2'  (rightmost plot) and which pixels, instead, moves some probability mass from `8' to `5'. 

The second and third rows of Figure \ref{fig:images} show attributions for the output of a ResNet-50 \citep{he2015deep} trained on ImageNet \citep{deng2009imagenet}.
We divide the image into a 32x32 multi-channel grid and collapse all but the 25 most probable classes to one (denoted `other').
The second row presents the case of a cat (classified as $c=$`Egyptian cat' by the model): the leftmost plot shows pixel importance as defined in Eq. \eqref{eq:no-change}. 
More interestingly, the three central figures show transition probabilities toward $c$ that highlight very different regions of the image: ears and tail for `Fox squirrel', a paw for `Plastic Bag` and portions of the face for the terrier.   
The right-most plot shows the pixel-wise most important (MAP) transitions (after thresholding) which may serve as a quick overview of the CSV. The third row shows additional examples of wrongly classified images: `Shoe shop' instead of `Confectionery' and `Monitor' rather than `Desktop computer'. 
In the latter case we see that the model (understandably) picks at the base of the screen as a distinguishing factor, while in the `Confectionery' example the model seems to mistake bottles  for shoes (in the background of the image). 
Appendix \ref{sec:fid} presents fidelity studies for these models aimed at corroborate the contrastive capabilities of the categorical values.

% When both players participate to the game, the logits are computed over the true image. When only one player participates, the pixels corresponding to the other player are set to 0, and the logits are computed over the resulting image. With no players, the logits are computed over a fully black image. The first row of Figure \ref{fig:mnist} shows a single digit, where we create several patches of pixels, progressively including more of the top part of the digit. When the patch is large enough, the model predicts 1 without the patch and 7 with it. 
% The categorical Shapley values, depicted in the second row through the full joint distribution $Q_i$,  
% % second row 
% show the average probability that adding the patch changes the prediction from any class to any other class. 
% In particular, we see that when the patch does not cover much of the top part of the digit, 
% the largest probability is in the diagonal, indicating no change.
% As the patch size increases, the largest probability occurs on the entry indicating transition from the label ``1'' to the label ``7``, with some remaining mass entering in ``2''.
% Compared to standard Shapley values, structured Shapley values are more informative, as they do not exclusively tell the importance of the patch for the prediction, but they can precisely determine the probability of the patch to change the prediction from a ``1'' to a ``7''.

Finally, we showcase an application to explain conditional probabilities of autoregressive LMs. 
We set up an experiment similar in spirit to FlipTest \citep{black2020fliptest} taking inspiration from \citep{nangia2020crows} where we probe the model for gender stereotyping on different sentences using ChatGPT-generated rephrases \citep{chatgpt2023}.
We compute average categorical differences between output given prompts with female versus male subject.
We restrict the output to a number of tokens in the order of 100 (depending on the sentence), picking a mix of manually selected, most probable (for a GPT2 model) and ChatGPT generated  short continuations.
We refer the reader to the appendix for details and for the formal definition of this experimental setting in a GT-XAI context.
We show results in Table \ref{tab:llm} for two types of sentences and two sizes of GPT2 models \citep{radford2019language}. 
Beside providing evidence of (known) stereotyping behaviour, we  
see
how CSV may offer fine-grained information about where precisely the probability mass moves, quantifying the probability that the change of the  sex of the subject  flips the predictions from e.g. `Lawyer' to `Nurse` in the job case.

\section{Conclusions}
\label{sec:conc}

\textbf{Related work in XAI.} 
Other XAI dimensions which we did not touch upon include: model-agnostic~\citep[inter alia]{ribeiro2016should} and model-specific~\citep{simonyan2013deep} , post-hoc~\citep{ribeiro2016should} and by design~\citep{alvarez2018towards}; distillation-based~\citep{tan2018distill}, feature-based~\citep{ribeiro2016should}, concept-based~\citep{kim2018interpretability}, and example-based~\citep{koh2017understanding}. See ~\citep{guidotti2018survey,arrieta2020explainable,gilpin2018explaining} for surveys and detailed list of methods. 
Several works~\citep{vstrumbelj2014explaining, lundberg2017unified,sundararajan2017axiomatic,sundararajan2020many,frye2020asymmetric} fall in the GT-XAI framework. The framework has also been explored for generating explanations in diverse contexts -- other than for local explanations -- see e.g. \citet{covert2020understanding, ghorbani2019data, ghorbani2020neuron} and \citet{mosca2022shap} for a survey.
% Shapley value based explanations are the most popular explainability method according to a recent study by~\citep{bhatt2020explainable}.
% 
% Several explainability techniques, Shapley values among them, have been criticized for their lack of infromativeness. A major reason for this criticism is that explanations from these techniques cannot be used to make contrastive statements which is a desired property in how humans parse explanations~\citep{miller2019explanation,kumar2020problems}. This led to conception of counterfactual explanations \citep{verma2020counterfactual} that answer the question the model would have generated a different output had values of features x and y been different. While counterfactual explanation identify features that maximally change the output (subject to some constraints), our framework makes contrastive statements (e.g., how much probability mass does a feature moves between different classes).
A very recent work proposes a technique to calibrate gradient based explanations of multi-class classifiers following a contrastive view \citep{wang2022not}. 
\citet{jacovi2021contrastive} propose a method for producing contrastive explanations (CEM) unrelated to the GT-XAI framework, while \citet{bowen2020generalized} discuss adaptations of SHAP to produce contrastive  explanations by formulating custom games. 
These formulations are linked to our categorical values. However distributional values offer a full probabilistic treatment and benefit from several theoretical properties as we showed in Section \ref{sec:properties}.
Recently, \citet{marx2023but} propose to aggregate explanations from multiple samples of a Gaussian process to construct uncertainty sets, which is loosely to our proposed sampling approach for estimating distributional values. 
In a concurrent work closely related to ours, ~\citet{chau2024explaining} focus on Gaussian processes and propose a stochastic Shapley value operator (GP-SHAP) whose output are multivariate Gaussians. They showcase its application to uncertainty quantification in explanations and predictive explanations.
Our work differs from GP-SHAP in two key aspects: we introduce a dependence structure among payoffs rather than assuming independence, and we take a distributional view instead of summing marginal contributions over coalitions.
Finally, from an application standpoint, in a study with human participants  \citet{fel2022cannot} identified a major concern for XAI technquies in their inability to reason about what the model is looking at. CSVs may offer answers to these types of questions due to their fine grain.

% \vspace{-1mm}
\textbf{Related work in CGT.} 
The Shapley value of simple games (i.e. games with payoffs in $\{0, 1\}$) has a probabilistic interpretation \citep[][pag. 168]{peleg2007introduction} however simple games are not stochastic. An ``and-or axiom" substitutes the linear axiom in simple games \citep{weber1988probabilistic}, here we extend to probabilistic  combinations.
% 
% \paragraph{Game theory.} The Shapley value of simple game has a probabilistic interpretation [intro to cop gt, p 168] however simple games are not probabilistic games. There's also a modified linearity for simple games [weber prob games; look for source]
% 
Extensions on the "domain" side, e.g. mulitlinear games \citep{owen1972multilinear}, regard games that are no longer defined on sets but on unit hypercubes.
In CGT, probabilistic games are typically intended as multi-stage games where the transition between stages is stochastic \citep{shapley1953stochastic, petrosjan2006cooperative} and not their intrinsic payoffs. 
% , [therefore not much related]. 
Static cooperative games with stochastic payoffs have been considered 
% \citep{charnes, SUJIS-BORM}
from the perspective of coalition formation and considering notions of players' utility \citep[e.g.][]{suijs1999cooperative} or studying two stages setups -- before and after the realisation of the payoff \citep[e.g.][]{granot1977cooperative}, and from an optimization perspective \citep{sun2022optimization}. 
To the best of our knowledge, our settings and constructions have not been studied before.

\textbf{Limitations.}  \label{sec:limitations} 
In this work, we have not touched upon several other (known) limitations of GT-XAI. 
Among these, two major issues are the computational complexity and the difficulty of defining meaningful  behaviour for out-of-coalition players. 
Regarding the first, we note that our value operators, being more informative than the standard counterparts, add (polynomial) computational cost, which is anyway overshadowed by the exponential cost of traversing coalitions. 
Integrating techniques for improved sampling recently proposed by \citet{mitchell2022sampling} may prove invaluable for the estimation of distributional values.
Regarding out-of-coalition behaviour, we have used in experiments a simple reference (or background) strategy, but note that many other formulations  \citep[e.g.][]{frye2020shapley, ren2023can} are possible. These are orthogonal dimensions to our  work. 

\textbf{Wrap up.} We have presented a framework that generalises the Shapley and related value operators for explaining more closely models with probabilistic outputs. 
Going forward, we believe that the same methodological approach -- i.e. reconsidering the way we formulate games and, by consequence, how we compute marginal contributions -- may be applied also to other contexts such as explaining spaces of functions or graphs (e.g. in causal discovery).
Another interesting direction of future research is to reconsider the type of payoff dependency we studied in this paper.
Finally, from a CGT perspective, we established a strong link to classic approaches and some other initial properties, such as efficiency and symmetry. We plan to continue the study, especially in the perspective of establishing contextually meaningfully properties with direct bearing in XAI.

% \clearpage

\section*{Impact statement}

Although this work is primarily concerned with the derivation and study of a novel class of value operators for cooperative stochastic games, we believe application to XAI may have generally a positive societal impact as they allow for greater scrutiny of model behaviour. 
As distributional values are in effect a strict extension of traditional approaches in game-theoretic XAI, we trust that their adoption may bring several benefits over using standard techniques such as SHAP and related (see also Appendix \ref{sec:adult} for a concrete example). 
However, we also acknowledge that misuse of explanatory techniques may potentially lead to miscalibration of stakeholders trust and the more complex technique introduced in this work may carry higher risks.
We therefore wish to highlight that at this stage of development distributional values are not meant as a ready-made XAI solution for the general public but should rather be applied and analysed by knowledgeable users.
% and are at present not meant to be XAI solution for the general public. 
In this sense, we intend to develop best practices for communication and visualization of the distributional values as well as continue probing the technique for failure cases and misinterpretations.

\section*{Acknowledgments}
We thank Cemre Zor, Ilja Kuzborskij, Gianluca Detommaso, Bilal Zafar, Camilla Damian, Sanjiv Das and Lukas Balles for helpful discussions and valuable feedback on this work.

\bibliography{dist_vals}
\bibliographystyle{icml2024}

%%%%%%%%%%%%%%%%%%%%%%%%%%%%%%%%%%%%%%%%%%%%%%%%%%%%%%%%%%%%%%%%%%%%%%%%%%%%%%%
%%%%%%%%%%%%%%%%%%%%%%%%%%%%%%%%%%%%%%%%%%%%%%%%%%%%%%%%%%%%%%%%%%%%%%%%%%%%%%%
% APPENDIX
%%%%%%%%%%%%%%%%%%%%%%%%%%%%%%%%%%%%%%%%%%%%%%%%%%%%%%%%%%%%%%%%%%%%%%%%%%%%%%%
%%%%%%%%%%%%%%%%%%%%%%%%%%%%%%%%%%%%%%%%%%%%%%%%%%%%%%%%%%%%%%%%%%%%%%%%%%%%%%%
\newpage
\appendix
\onecolumn

\section{On the dependency  structure of the payoffs}

In this work, for the reasons outlined in Section \ref{sec:dist-vals}, we propose a simple and natural dependency structure between all the payoffs of a game in that $v(S)=v(S, \varepsilon)$, using a deterministic reparameterization and ``noise sharing'' of $\varepsilon\sim \rho(\varepsilon)$.
The stochasticity of the output is captured by $\varepsilon$ while the difference due to the coalition on which $v$ is computed is captured through $g$. 
This dual treatment fits our typical context where (single) ML models underlie our structured games:  different inputs correspond to different coalitions, with their variation encoded by $f$. 
However, the model itself (e.g. the parameters of a neural net) remains unchanged across evaluations of different inputs (i.e. coalitions). 
We encoded by the presence of a shared source of randomness.
% In operative terms, this corresponds to fixing a random seed for all the evaluations  
From an operational standpoint, as we note in the first footnote of Section \ref{sec:analytic-exp} we may also interpret the $\varepsilon\sim \rho(\varepsilon)$ as the ``random seed'' that we use to compute the output of a model. 
Then, from this perspective, sharing randomness corresponds to fixing a random seed for all evaluations. 
% This ensures that 

Another justification has already been discussed in the main body, contrasting this choice with the possibility of assuming (full) independence between the various payoffs:
independence would lead to marginal contributions being non-zero (in the probabilistic sense) even when the  parameters of the probability distributions $v(S\cup i)$ and $v(S)$ would be the same (e.g. same success probability, in the Bernoulli case). 
However, given a latent variable representation of the marginal distribution of interest there are other dependency assumptions we could explore that may also better capture underlying stochasticiy in the model (e.g. Bayesian nets). Although we do not cover these cases in the current presentation, we offer next some possible direction in this sense. 

\textbf{Exchangeability.} 
A weaker assumption on the variables $v(S)$ would be {\em exchangeability}: for any subset $S_1, \dots, S_k$ and any permutation $\pi(j)$, the joint distributions of $(v(S_1), \dots, v(S_k))$ and $(v(S_{\pi(1)}), \dots, v(S_{\pi(k)}))$ are the same. By de Finetti's theorem~\citep{Hewitt:55}, there exists a shared random variable $\mathbf{\varepsilon}$ so that the $v(S)$ become independent when we condition on $\mathbf{\varepsilon}$. This is weaker than our assumption of determinism given $\mathbf{\varepsilon}$, since each $v(S)$ can still have independent randomness given $\mathbf{\varepsilon}$. Studying games with random payoffs under a weaker exchangeability assumption is an interesting topic for further research.
% \end{remark}

\section{Derivation of the analytical  expressions of the Categorical values} 
% Structured Values for Categorical Games}
\label{app:value-categorical}

We provide a derivation of the expressions $\tilde{Q}_{i,S}(r, s)$ in Section~\ref{sec:categorical-games}, paragraph ``Categorical values''. In this derivation, $i$ and $S$ are fixed, and we write $\mathcal{P}_{r s}$ for $\tilde{Q}_{i,S}(r, s)$. Let $d\ge 3$ be an integer, $[\alpha_j]$ and $[\beta_j]$ be sets of $d$ real numbers. Above, $\alpha_j = \theta_{S\cup i, j}$ and $\beta_j = \theta_{S, j}$, but the derivation below does not make use of this. Also, let $\varepsilon_j$ be $d$ independent standard Gumbel variables, each of which has distribution function and density
\[
  F(\varepsilon) = \exp\left( e^{-\varepsilon} \right),\quad p(\varepsilon) = F(\varepsilon)' =
  \exp\left( -\varepsilon - e^{-\varepsilon} \right) = e^{-\varepsilon} F(\varepsilon).
\]
Fix $r, s\in\srng{d}$, $r\ne s$. We would like to obtain an expression for the probability $\mathcal{P}_{r s}$ of
\[
  \argmax_j\left( \alpha_j + \varepsilon_j \right) = r\quad\textrm{and}\quad
  \argmax_j\left( \beta_j + \varepsilon_j \right) = s.
\]
Define
\[
  \alpha_{j r} := \alpha_j - \alpha_r,\quad \beta_{j s} := \beta_j - \beta_s.
\]
The $\argmax$ equalities above can also be written as a set of $2 d$ inequalities (2 of which are trivial):
\[
  \varepsilon_j \le \varepsilon_r - \alpha_{j r},\quad \varepsilon_j\le \varepsilon_s - \beta_{j s},\quad
  j=\rng{d}.
\]
Then:
\[
  \mathcal{P}_{r s} = \Ex\left[ \prod\nolimits_j I_j \right],\quad I_j :=
   \Ind{\varepsilon_j \le \min(\varepsilon_r - \alpha_{j r}, \varepsilon_s - \beta_{j s})}.
\]
Two of them are simple:
\[
  I_r = \Ind{\varepsilon_r\le \varepsilon_s - \beta_{r s}},\quad I_s = \Ind{\varepsilon_s\le \varepsilon_r -
    \alpha_{s r}}, \quad I_r I_s = \Ind{\alpha_s - \alpha_r \le \varepsilon_r- \varepsilon_s \le
    \beta_s - \beta_r}.
\]
Denote
\[
  \gamma_j := \alpha_{j r} - \beta_{j s} = \nu_j - (\alpha_r - \beta_s),\quad
  \nu_j := \alpha_j - \beta_j.
\]
Note that $\gamma_j$ depends on $r, s$, but $\nu_j$ does not. If $j\ne r, s$, then
\[
  I_j = \Ind{\varepsilon_j\le \varepsilon_r - \alpha_{j r}} \Ind{\varepsilon_r - \varepsilon_s \le \gamma_j}
  + \Ind{\varepsilon_j\le \varepsilon_s - \beta_{j s}} \Ind{\varepsilon_r - \varepsilon_s \ge \gamma_j}.
\]
If we exchange sum and product, we obtain an expression of $\mathcal{P}_{r s}$ as sum of $2^{d - 2}$ terms. Each of these terms is an expectation over $\varepsilon_r$, $\varepsilon_s$, with the argument being the product of $d-2$ terms $F(\varepsilon_r + a_j)$ or $F(\varepsilon_s + a_j)$ and a box indicator for $\varepsilon_r - \varepsilon_s$. In the sequel, we make this more concrete and show that at most $d-1$ of these terms are nonzero.

With a bit of hindsight, we assume that $\nu_1\ge \nu_2\ge \dots\ge \nu_d$, which is obtained by reordering the categories. This implies that $[\gamma_j]$ is nonincreasing for all $(r, s)$. Also, define the function $\pi(k) = k + \Ind{r\le k} + \Ind{s-1\le k}$ from $\srng{d-2}$ to $\srng{d}\setminus\{r, s\}$. We will argue in terms of a recursive computation over $k=\rng{d-2}$. Define
\[
  M_k(\varepsilon_r, \varepsilon_s) = \Ex\left[ I_r I_s \prod\nolimits_{1\le j\le k} I_{\pi(j)}\; \bigl|\;
    \varepsilon_r, \varepsilon_s \right],\quad k\ge 0,
\]
so that $\mathcal{P}_{r s} = \Ex[M_{d-2}(\varepsilon_r, \varepsilon_s)]$. Each $M_k$ can be written as sum of $2^{k}$ terms. Imagine a binary tree of depth $d-1$, with layers indexed by $k=0, 1, \dots, d-2$. Each node in this tree is annotated by a box indicator for $\varepsilon_r - \varepsilon_s$ and some information detailed below. We are interested in the $2^{d-2}$ leaf nodes of this tree.

\subsection{Box indicators. Which terms are needed?}

We begin with a recursive computation of the box indicators, noting that we can eliminate all nodes where the box is empty. Label the root node (at $k=0$) by $1$, its children (at $k=1$) by $10$ (left), $11$ (right), and so on, and define the box indicators as $\Ind{l_1\le \varepsilon_r - \varepsilon_s\le u_1}$, and $(l_{10}, u_{10})$, $(l_{11}, u_{11})$ respectively. Then, $l_1 = \alpha_s - \alpha_r$, $u_1 = \beta_s - \beta_r$ defines the box for the root. Here,
\[
  l_1 \ge u_1\quad\Leftrightarrow\quad \nu_s \ge \nu_r.
\]
Since $[\nu_j]$ is non-increasing, the root box is empty if $s < r$, so that $\mathcal{P}_{r s} = 0$ in this case. In the sequel, we assume that $r < s$ and $\nu_r > \nu_s$, so that $l_1 < u_1$.

If $\vn{}$ is the label of a node at level $k - 1$ with box $(l_{\vn{}}, u_{\vn{}})$, then
\[
  l_{\vn{} 0} = l_{\vn{}},\quad u_{\vn{} 0} = \min(\gamma_{\pi(k)}, u_{\vn{}}),\quad
  l_{\vn{} 1} = \max(\gamma_{\pi(k)}, l_{\vn{}}),\quad  u_{\vn{} 1} = u_{\vn{}}.
\]
Consider node $11$ (right child of root). There are two cases. (1) $\gamma_{\pi(1)} < u_1$. Then, $l_{11} \ge \gamma_{\pi(1)}\ge \gamma_{\pi(k)}$ for all $k\ge 1$, so all descendants must have the same $l = l_{11}$. If ever we step to the left from here, $u = \min(\gamma_{\pi(k)}, u_1)\le \gamma_{\pi(k)}\le \gamma_{\pi(1)}\le l_{11}$, so the node is eliminated. This means from $11$, we only step to the right: $111, 1111, \dots$, with $l = \max(\gamma_{\pi(1)}, l_1)$, $u = u_1$, so there is only one leaf node which is a descendant of $11$. (2) $\gamma_{\pi(1)}\ge u_1$. Then, $l_{11} \ge u_{11}$, so that $11$ and all its descendants are eliminated.

At node $10$, we have $l_{10} = l_1$. If $\gamma_{\pi(1)}\le l_1$, the node is eliminated, so assume $\gamma_{\pi(1)} > l_1$, and $u_{10} = \min(\gamma_{\pi(1)}, u_1)$. Consider its right child $101$. We can repeat the argument above. There is at most one leaf node below $101$, with $l = \max(\gamma_{\pi(2)}, l_1)$ and $u = u_{10} = \min(\gamma_{\pi(1)}, u_1)$.

All in all, at most $d-1$ leaf nodes are not eliminated, namely those with labels $1 0\dots 0 1 \dots 1$, and their boxes are
$[\max(\gamma_{\pi(1)}, l_1), u_1]$, $[\max(\gamma_{\pi(2)}, l_1), \min(\gamma_{\pi(1)}, u_1)]$, $\dots$, $[\max(\gamma_{\pi(d-2)}, l_1), \min(\gamma_{\pi(d-3)}, u_1)]$, $[l_1, \min(\gamma_{\pi(d-2)}, u_1)]$.

Recall that each node term is a product of $d-2$ Gumbel CDFs times a box indicator. What are these products for our $d-1$ non-eliminated leaf nodes? The first is
$F(\varepsilon_s - \beta_{\pi(1) s}) \cdots F(\varepsilon_s - \beta_{\pi(d-2) s})$, the second is $F(\varepsilon_r - \alpha_{\pi(1) r}) F(\varepsilon_s - \beta_{\pi(2) s}) \cdots F(\varepsilon_s - \beta_{\pi(d-2) s})$, the third is $F(\varepsilon_r - \alpha_{\pi(1) r}) F(\varepsilon_r - \alpha_{\pi(2) r}) F(\varepsilon_s - \beta_{\pi(3) s}) \cdots F(\varepsilon_s - \beta_{\pi(d-2) s})$ and the last one is $F(\varepsilon_r - \alpha_{\pi(1) r}) \cdots F(\varepsilon_r - \alpha_{\pi(d-2) r})$. Next, we derive expressions for the expectation of these terms.

\subsection{Analytical expressions for expectations}

Consider $d-2$ scalars $a_1, \dots, a_{d-2}$ and $1\le k\le d-1$. We would like to compute
\begin{equation}\label{eq:def-big-a}
  A = \Ex\left[ \left( \prod\nolimits_{j < k} F(\varepsilon_r + a_j) \right)
    \left( \prod\nolimits_{j \ge k} F(\varepsilon_s + a_j) \right)
    \Ind{l\le \varepsilon_r - \varepsilon_s\le u} \right].
\end{equation}
Denote
\[
  G(a_1, \dots, a_t) := \Ex[F(\varepsilon_1 + a_1) \cdots F(\varepsilon_1 + a_t)].
\]
We start with showing that
\[
  G(a_1, \dots, a_t) = \left( 1 + e^{-a_1} + \cdots + e^{-a_t} \right)^{-1}.
\]
Recall that $p(x) = F(x)' = e^{-x} F(x)$. If $\tilde{F}(x) = \prod_{j=1}^{t} F(x + a_j)$, then
\[
  \tilde{F}(x)' = \left( \sum\nolimits_{j=1}^{t} e^{-a_j} \right) e^{-x} \tilde{F}(x).
\]
Using integration by parts:
\[
  G(a_1, \dots, a_t) = \int \tilde{F}(x) p(x)\, d x = 1 - \int \tilde{F}(x)' F(x)\, d x
  = 1 - \left( \sum\nolimits_{j=1}^{t} e^{-a_j} \right) G(a_1, \dots, a_t),
\]
where we used that $F(x) = e^x p(x)$.

Next, define
\[
  g_1 = \log\left( 1 + e^{-a_1} + \cdots + e^{-a_{k-1}} \right),\quad
  g_2 = \log\left( 1 + e^{-a_{k}} + \cdots + e^{-a_{d-2}} \right).
\]
We show that $A$ in \eqp{def-big-a} can be written in terms of $(g_1, g_2, l, u)$ only. Assume that $k>1$ for now. Fix $\varepsilon_s$ and do the expectation over $\varepsilon_r$. Note that $\Ind{l\le \varepsilon_r - \varepsilon_s\le u} = \Ind{\varepsilon_s + l\le \varepsilon_r\le \varepsilon_s + u}$. If $\tilde{F}(x) = \prod_{j<k} F(x + a_j)$, then
\[
  \tilde{F}(x)' = \left( \sum\nolimits_{j<k} e^{-a_j} \right) e^{-x} \tilde{F}(x).
\]
Using integration by parts:
\[
  B(\varepsilon_s) = \int_{\varepsilon_s + l}^{\varepsilon_s + u} \tilde{F}(x) p(x)\, d x =
    \left[ \tilde{F}(x) F(x) \right]_{\varepsilon_s + l}^{\varepsilon_s + u} - B(\varepsilon_s)
    \sum_{j<k} e^{-a_j},
\]
so that
\[
  B(\varepsilon_s) = e^{-g_1} \left[ \tilde{F}(x) F(x) \right]_{\varepsilon_s + l}^{\varepsilon_s + u}
\]
and
\[
  A = \Ex\left[ B(\varepsilon_s) \prod\nolimits_{j\ge k} F(\varepsilon_s + a_j) \right]
  = A_1 - A_2,
\]
where
\[
\begin{split}
  A_1 & = e^{-g_1} \Ex\left[ \left( \prod\nolimits_{j<k} F(\varepsilon_s + u + a_j)
    \right) \left( \prod\nolimits_{j\ge k} F(\varepsilon_s + a_j) \right)
    F(\varepsilon_s + u) \right] \\
  & = e^{-g_1} G(a_1 + u, a_2 + u, \dots, a_{k-1} + u, a_{k}, \dots, a_{d-2}, u)
\end{split}
\]
and
\[
  A_2 = e^{-g_1} G(a_1 + l, a_2 + l, \dots, a_{k-1} + l, a_{k}, \dots, a_{d-2}, l).
\]
Now,
\[
\begin{split}
  -\log A_1 & = g_1 - \log G(a_1 + u, a_2 + u, \dots, a_{k-1} + u, a_{k}, \dots,
  a_{d-2}, u) \\
  & = g_1 + \log\left( 1 + \sum\nolimits_{j<k} e^{-a_j - u} +
  \sum\nolimits_{j\ge k} e^{-a_j} + e^{-u} \right) = g_1 + \log\left(
  e^{g_2} + e^{-u + g_1} \right) \\
  & = g_1 + g_2 + \log\left( 1 + e^{g_1 - g_2 - u} \right)
\end{split}
\]
and
\[
  -\log A_2 = g_1 + g_2 + \log\left( 1 + e^{g_1 - g_2 - l} \right)
\]
so that
\begin{equation}\label{eq:res-big-a}
  A = A_1 - A_2 = e^{-(g_1 + g_2)}\left( \sigma(g_2 - g_1 + u) - \sigma(g_2 - g_1 + l)
  \right),\quad \sigma(x) := \frac{1}{1 + e^{-x}}.
\end{equation}
If $k=1$, we can flip the roles of $\varepsilon_r$ and $\varepsilon_s$ by $g_1\leftrightarrow g_2$, $l\to -u$, $u\to -l$, $k\to d-1$, which gives
\[
  e^{-(g_1 + g_2)}\left( \sigma(-(g_2 - g_1 + l)) - \sigma(-(g_2 - g_1 + u))
  \right) = e^{-(g_1 + g_2)}\left( \sigma(g_2 - g_1 + u) - \sigma(g_2 - g_1 + l)
  \right),
\]
using $\sigma(-x) = 1 - \sigma(x)$, so the expression holds in this case as well.

\subsection{Efficient computation for all pairs}

Our $d-1$ terms of interest can be indexed by $k=\rng{d-1}$. We can use the analytical expression just given with $a_j = -\alpha_{\pi(j) r}$ for $1\le j< k$ and $a_j = -\beta_{\pi(j) s}$ for $k\le j\le d-2$. Define
\[
  g_1(k) = \log\left( 1 + \sum\nolimits_{1\le j < k}
    e^{\alpha_{\pi(j)} - \alpha_r} \right),\quad
  g_2(k) = \log\left( 1 + \sum\nolimits_{k \le j\le d-2}
    e^{\beta_{\pi(j)} - \beta_s} \right),
\]
as well as
\[
  l(k) = \max(\gamma_{\pi(k)}, l_1),\quad u(k) = \min(\gamma_{\pi(k-1)}, u_1),
\]
where we define $\pi(0) = 0$, $\pi(d-1) = d+1$, $\gamma_0 = +\infty$, and $\gamma_{d+1} = -\infty$. Note that
\begin{equation}\label{eq:bounds-l-u}
\begin{split}
  & l(k) = \max( \nu_{\pi(k)} - \alpha_r + \beta_s, \alpha_s - \alpha_r ) =
  \beta_s - \alpha_r + \max( \nu_{\pi(k)}, \nu_s ), \\
  & u(k) = \min( \nu_{\pi(k-1)} - \alpha_r + \beta_s, \beta_s - \beta_r ) =
  \beta_s - \alpha_r + \min( \nu_{\pi(k-1)}, \nu_r ).
\end{split}
\end{equation}
$\mathcal{P}_{r s}$ is obtained as sum of $A(g_1(k), g_2(k), l(k), u(k))$ for $k=\rng{d-1}$. In the sequel, we show how to compute these terms efficiently, for all pairs $r < s$.

Recall that $\gamma_j = \nu_j - (\alpha_r - \beta_s)$, $u_1 = \beta_s - \beta_r$, $l_1 = \alpha_s - \alpha_r$. Then:
\[
  l(k) < u(k)\quad \Leftrightarrow\quad \nu_{\pi(k)} < \nu_{\pi(k-1)}\;
  \wedge\;  \nu_{\pi(k)} < \nu_r\; \wedge\; \nu_s < \nu_{\pi(k-1)}.
\]
Recall that $\pi(k) = k + \Ind{r\le k} + \Ind{s-1\le k}$. Define $K_1 = \srng{r-1}$, $K_3 = \srng[s]{d-1}$, each of which can be empty. For $k\in K_1$, $\nu_{\pi(k)} = \nu_k \ge \nu_r$, so $l(k)\ge u(k)$. For $k\in K_3$, we have $\pi(k-1) = k+1 > s$, so that $\nu_s\ge \nu_{\pi(k-1)}$ and $l(k)\ge u(k)$. This means we only need to iterate over $k\in K_2 = \srng[r]{s-2}$ with $\pi(k) = k+1$ and $k=s-1$ with $\pi(k) = s+1$ (the latter only if $s < d$).

As $k$ runs in $K_2$, $\pi(k) = r+1,\dots, s-1$, and if $s < d$ then $\pi(s-1) = s+1$. Now
\[
  g_1(k) = \log\left( 1 + \sum\nolimits_{1\le j < k}
    e^{\alpha_{\pi(j)} - \alpha_r} \right) = \log\sum\nolimits_{1\le j\le k}
    e^{\alpha_{j} - \alpha_r},
\]
using that $e^{\alpha_r - \alpha_r} = 1$. For $g_2(k)$, if $k < s-1$, then $\{\pi(j)\, |\, k\le j\le d-2\} = \srng[k+1]{d}\setminus\{s\}$, and if $k = s-1$, the same holds true (the set is empty if $s=d$). Using $e^{\beta_s - \beta_s} = 1$, we have
\[
  g_2(k) = \log\sum\nolimits_{k < j\le d} e^{\beta_{j} - \beta_s}.
\]
Define
\[
  \bsalpha{k} := \log\sum_{j=1}^k e^{\alpha_j},\quad \bsbeta{k} :=
  \log\sum_{j=k+1}^d e^{\beta_j},\quad k=\rng{d-1}.
\]
Then:
\[
  g_1(k) = \bsalpha{k} - \alpha_r,\quad g_2(k) = \bsbeta{k} - \beta_s,\quad
  k = \rng[r]{s-1}.
\]
Finally, using $g_2(k) - g_1(k) = \bsbeta{k} - \bsalpha{k} + \alpha_r - \beta_s$ and \eqp{bounds-l-u}, we have
\[
  g_2(k) - g_1(k) + l(k) = \bsbeta{k} - \bsalpha{k} + \max( \nu_{\pi(k)}, \nu_s ),, \quad
  g_2(k) - g_1(k) + u(k) = \bsbeta{k} - \bsalpha{k} + \min( \nu_{\pi(k-1)}, \nu_r ).
\]
Some extra derivation, distinguishing between (a) $r = s-1$, (b) $r < s-1 \wedge k\in K_2$, (c) $r < s-1 \wedge k = s-1$ shows that
\[
  \max( \nu_{\pi(k)}, \nu_s ) = \nu_{k+1},\quad
  \min( \nu_{\pi(k-1)}, \nu_r ) = \nu_k,\quad k=\rng[r]{s-1}.
\]
Plugging this into \eqp{res-big-a}:
\[
  A(k) = e^{\alpha_r + \beta_s} c_k,\quad c_k =
  e^{-\bsbeta{k} - \bsalpha{k}} \left( \sigma\left( \bsbeta{k} - \bsalpha{k}
      + \nu_{k} \right) - \sigma\left( \bsbeta{k} - \bsalpha{k} + \nu_{k+1}
    \right) \right).
\]
and $\mathcal{P}_{r s} = \sum_{k=r}^{s-1} A(k)$. Importantly, $c_k$ does not depend on $r, s$. Therefore:
\begin{equation}\label{eq:probability-r-s}
  \mathcal{P}_{r s} = e^{\alpha_r + \beta_s} (C_s - C_r),\quad C_t =
  \sum_{k=1}^{t-1} c_k\quad (r < s);\quad \mathcal{P}_{r s} = 0\quad (r > s).
\end{equation}
The sequences $[\bsalpha{k}]$, $[\bsbeta{k}]$, $[c_k]$ , $[C_k]$ can be computed in $\mathcal{O}(d)$.

Finally, we also determine $\mathcal{P}_{r r}$, which is defined by the inequalities $\varepsilon_j \le \varepsilon_1 - \max(\alpha_{j r}, \beta_{j r})$. A derivation like above (but simpler) gives:
\[
  \mathcal{P}_{r r} = \left( 1 + \sum_{j\ne r} e^{\max(\alpha_{j r}, \beta_{j r})} \right)^{-1}.
\]
Now, $\alpha_{j r}\ge \beta_{j r}$ iff $\nu_j\ge \nu_r$ iff $j < r$, so that
\[
\begin{split}
  \mathcal{P}_{r r} & = \left( 1 + \sum_{j<r} e^{\alpha_j - \alpha_r} + \sum_{j>r}
    e^{\beta_j - \beta_r} \right)^{-1} = \left( e^{\bsalpha{r} - \alpha_r} +
    e^{\bsbeta{r} - \beta_r} \right)^{-1} \\
    & = e^{\beta_r -\bsbeta{r}}
  \sigma(\bsbeta{r} - \bsalpha{r} + \nu_r),\quad (r < d), \\
  \mathcal{P}_{d d} & = e^{\alpha_d - \bsalpha{d}}.
\end{split}
\]

\section{Extended background and proof of Proposition \ref{prop:all}}

In this section we extend the background on cooperative game theory of Section \ref{sec:pre} and then provide a proof for the Proposition \ref{prop:all}.

Our definition of distributional values depend on the coalition structure $\{p^i\}=p$ for $i=[n]$, where the $p^i$ are PMFs over coalitions, one for each player.
This formulation inherits from multiple generalisations of the Shapley value appearing in CGT \citep{weber1988probabilistic, dubey1981value}, which comprises operators $\phi = (\phi_i)_{i=1}^n: \mathcal{G}_n \mapsto \mathbb{R}^n$ that may be written as expectations of marginal contributions $v(S\cup i) - v(S)$ as follows:
\begin{equation}
    \label{eq:prob_val-apx}
     \phi_i(v) = \sum_{S\in 2^{[n]\setminus i}} p^i(S) [v(S\cup i) - v(S)] 
    = \mathbb{E}_{S\sim p^i(S)} [v(S\cup i) - v(S)].
\end{equation}
Probabilistic (group) values, semivalues, random-order group values (also known as asymmetric Shapley values \citep{frye2020asymmetric}) and the Shaply value can be written in this way. 
Semivalues and random-order group values are probabilistic values and the Shapley value is the only operator that is both a semivalue and a random-order group value. 
Furthermore, one can think of random-order group values as originating from a single shared probability distribution over permutations (rather than coalitions) of players $\nu:\Pi_n\mapsto[0,1]$ as follows;
\begin{equation*}
    \label{eq:roval-apx}
    \phi_i(v) =  \sum_{\pi\in\Pi_n} \nu(\pi) [v(\{j\leq \pi(i)\}) - v(\{j< \pi(i)\})] =\mathbb{E}_{\pi \sim \nu(\pi)}[v(\{j\leq \pi(i)\}) - v(\{j< \pi(i)\}],
\end{equation*}
where $\Pi_n$ is the set of all permutations of $[n]$.
In this view, the Shapley value is the random order group value with uniform probability over permutations; i.e. $\nu(\pi)=(n!)^{-1}$.
% to the random 

\subsection{Axioms of the value operators}
The four classes of value operators are traditionally derived, studied, and presented in relation to a number of axioms that they satisfy. \footnote{In contrast, we adopt a constructive view in the main paper and speak about ``properties''.} 
We list the principal five axioms below in the context of standard real-valued games $v:2^{[n]}\mapsto \mathbb{R}$.
\paragraph{Dummy}
A player $i$ is a dummy for $v$ if for every $S\neq \emptyset$,
$v(S\cup i) = v(S) + v(i)$.
A value operator $\phi$ satisfies the dummy axiom if $\phi_i(v) = v(i)$ whenever a player $i$ is dummy for a game $v$.  
% \end{definition}

% All the four values
This axiom encompasses the null player axiom found e.g. in  \citep{lundberg2017unified} which can be obtained as special case when  $v(i)=0$. 
The dummy axiom essentially states that if a player has no strategic impact on the game, then it shall be  assigned exactly the payoff that it receives by playing alone.

\paragraph{Linearity}
Let $v = w + u$, meaning that $v(S)=w(S) + u(S)$ for all coalitions, 
where $v, w, u$ are all $n-$players games.
A value operator satisfies the linearity axiom if $\phi(v) = \phi(w) + \phi(u)$
% \end{definition}

This axiom essentially requires $\phi$  be a linear operator between the two vector spaces $\mathcal{G}_n$ and $\mathbb{R}^n$. 

\paragraph{Monotonicity.}
A game $v$ is monotonic if for every $S\subseteq T$ $v(S)\leq v(T)$. 
A value operator satisfies the monotonicity axiom if $\phi_i(v) \geq 0$ for all $i\in[n]$ whenever $v$ is a monotonic game.

% \end{definition}
This axiom requires that the values of players of monotonic games be positive and encodes the idea that for games that have non-decreasing payoffs for increasing coalition sizes, there can be no harm in joining a coalition. 

All the four classes of group values satisfy these three axioms. The following two axioms are instead only satisfied by random-order group values (efficiency) and semivalues (symmetry), respectively. 
The Shapley value satisfies both of them at the same time.

\paragraph{Efficiency.}
Let $v(\emptyset) = 0$. A value operator is efficient if $\sum_i \phi_i(v) = v([n])$. 
% \end{definition}

If $v(\emptyset)\neq 0$ then one can still talk about efficiency by subtracting the offset $v(\emptyset)$ from the grand payoff.
It can be shown \citep{weber1988probabilistic} that if the coalition distribution is such that 
\begin{equation}
    \label{eq:effi:a}
    \sum_{i\in[n]} p^i([n]\setminus i) = 1 \quad \text{and} \quad  \quad \sum_{i\in S} p^i(S\setminus i) = \sum_{j\not\in S} p^j(S)
\end{equation}
than the associated value operator is efficient. 
We refer to coalition structure that satisfy Eq. \eqref{eq:effi:a} as \textit{efficient}.
Coalition distribution deriving form random-order group values are efficient.

\paragraph{Symmetry}
A value operator is symmetric if for every permutation $\pi$ of $[n]$ $\phi_i(v) = \phi_{\pi(i)}(\pi v)$ where $\pi v$ is the game defined as $\pi v(\{\pi(i) \, : \, i \in S\}) = v(S)$. 

% As for 
% One can show that if an operaot
In particular, simmetry entails that if $i$ and $j$ are indistinguishable players for a game $v$, i.e. $v(S\cup i) = v(S\cup j)$ for all $S$, then $\phi_i(v) =\phi_j(v)$.
If an operator is symmetric, then the coalition PMFs are shared among players and only depend on the coalition dimension, i.e. there exist a PMF $\bar{p}$ over $[n-1]$ such that
\begin{equation}
    \label{eq:symm:a}
    p^i(S)=\bar{p}(|S|) \quad \text{for all } i\in[n], \, S\in 2^{[n]\setminus i}.
\end{equation} 
As we did for the efficiency case, we refer to coalition structure with this property as \textit{symmetric}.
Distributions deriving from semivalues are symmetric.

\subsection{Proof of Proposition \ref{prop:all}}

\proof 

\begin{enumerate}
    \item[$(i)$] For each $i$, by direct computation and linearity of the expectation, we have that
    \begin{align*}
        \mathbb{E}_{S, \varepsilon}[\xi_i(v)] &= 
        \mathbb{E}_{S\sim p^i(S)} [\mathbb{E}_{\varepsilon\sim p(\varepsilon)}[v(S\cup i, \varepsilon) - v(S, \varepsilon) ] \\
        &=\mathbb{E}_{S\sim p^i(S)} [\bar{v}(S\cup i) - \bar{v}(S) ] = \{\phi_i(\bar{v})\}_{i=1}^d, 
    \end{align*}
    where $d$ is the dimension of the output space. 
    Note that for the specific distributions we covered in Section \ref{sec:analytic-exp},  have $u(S) = \pi_S$ for Bernoulli games, $u(S)=\mu_S$ for Gaussian games, and $u(S) = \mathrm{Softmax}(\theta_S)$ for Categorical games.
    \item[$(ii)$] This is a direct consequence of the reparameterization condition we introduced in Section \ref{sec:dist-vals}.
    \item[$(iii)$] For every $z\in T$, we have that 
    \begin{equation*}
        q_i(z) = q_i(z | v=v')\mathbb{P}(v=v') + q_i(z|v=v'')\mathbb{P}(v=v'') = \pi q'_i(z) + (1-\pi) q''_i(z)
    \end{equation*}
    where $q'$ and $q''$ are probability distributions of the PSVs of $v'$ and $v''$, respectively.
    \item[$(iv)$] Recall that if a coalition distribution is efficient, then its PMF follows the conditions Eq. \eqref{eq:effi}. Then, we have
    \begin{align}
        \sum_{i\in[n]} \mathbb{E}_{S\sim p^i(S)}[\xi_i(v)] &= \sum_{i\in[n]} 
        \sum_{S\subseteq 2^{[n] \setminus i}} p^i(S)[v(S\cup i, \varepsilon) - v(S, \varepsilon)]
        \nonumber
        \\
        &= \sum_{S\in 2^{n} \setminus \{[n], \emptyset\}} v(S, \varepsilon) \left[
            \sum_{j\in S} p^j(S\setminus j) - \sum_{j\not\in S} p^j(S) 
                    \right]
                    \label{eq:p-rest}
            \\
            \label{eq:p-full}
            &+ v([n], \varepsilon) \sum_{i\in [n]} p^i([n] \setminus i) 
            \\
            &+ 
            \label{eq:p-empty}
            v(\emptyset, \varepsilon) \sum_{i \in [n]} p^i(\emptyset)
            \\
            &=  v([n], \varepsilon) - v(\emptyset, \varepsilon) = v([n]) \ominus v(\emptyset), \nonumber
    \end{align}
    where, because of the efficiency hypothesis, the difference of sums of probabilities in line \eqref{eq:p-rest} are zero, and the probabilities in line \eqref{eq:p-full} and line \eqref{eq:p-empty} sum both to one. 
    To see that the summation in Eq. \eqref{eq:p-empty} is one as a consequence of \eqref{eq:effi}, consider that for any $k\in [n-1]$ 
    \begin{align*}
        \sum_{|S|=k} \sum_{i\in S} p^i(S\setminus i) &= \sum_{|S|=k} \sum_{i\not\in S} p^i(S) = \sum_{|S|=k} \sum_{i\not\in S}  p^i((S \cup i) \setminus i) \\
        &= \sum_{|S|=k+1} \sum_{i\in S} p^i(S\setminus i),
    \end{align*}
    creating a chain of equalities and conclude by taking $k=1$ and $k'=n-1$. 
    \item[$(v)$]
    Assume $p$ satisfies Eq. \eqref{eq:symm}. We prove a the more general property of symmetry, let $\pi\in \Pi_n$ be a permutation of $[n]$ and define $\pi v$ as above. 
    Then 
    \begin{align*}
        \xi_i(v) &= v(S\cup i) \ominus v(S) = \pi v(\pi(S\cup i)) \ominus \pi v(\pi(S)) \quad \text{for } \varepsilon \sim \rho(\varepsilon), \; S \sim p^i(S)
        \\
        &= \xi_{\pi(i)}(\pi v)
    \end{align*}
    where the second equality holds because the probabilities of $S$ do not depend on the player $i$ and where we denote $\pi(S) = \{\pi(i) \, : \, i \in S\}$.
\end{enumerate}

\section{Further experimental details and results}
\label{sec:further}

We run all the experiments on a machine with 8 Intel(R) Xeon(R) Platinum 8259CL CPUs @ 2.50GHz  and one Nvidia(R) Tesla(R) V4 GPU. 
Python code is available at \url{https://github.com/amazon-science/explaining-probabilistic-models-with-distributinal-values}.
% upon acceptance. 

\subsection{Mnist}

\begin{figure}
    \centering
    \includegraphics[width=0.7\textwidth]{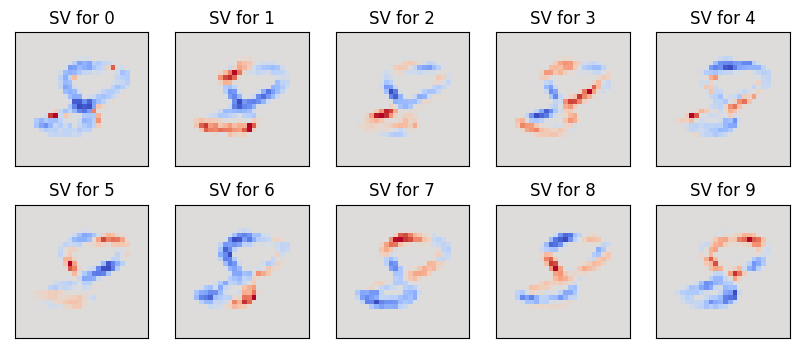}
    \includegraphics[width=0.7 
    \textwidth]{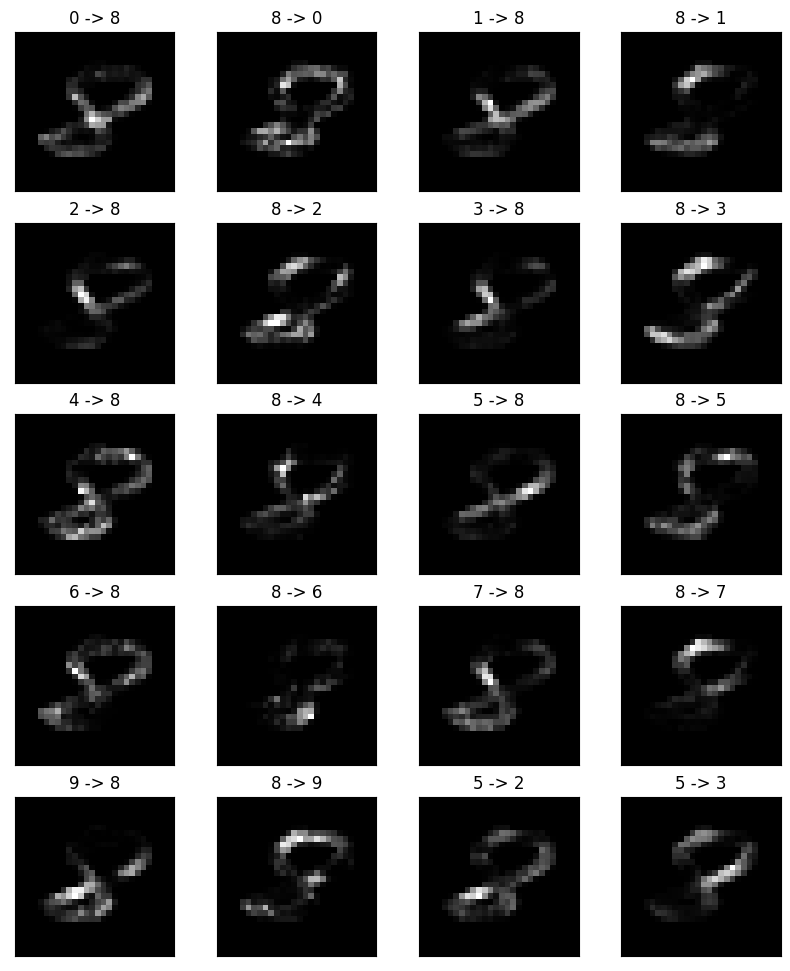}
    \begin{small}
    \caption{(Top two rows) We plot the standard (estimated) Shapley value for each of the digit explaining the output probabilities: red gradations indicate positive contribution, blue negative. 
    The values have been obtained as expectation of the Categorical SV, but could have been obtained also with other techniques such as KernelSHAP \citep{lundberg2017unified}. (Bottom five rows) We plot slices of the Categorical SV. 
    All plots except the last two show transition probabilities from and to the digit `8' and complement Figure \ref{fig:images} in the main paper. The last two plots show examples of transition probabilities that do not involve the digit `8'.}
    \label{fig:mnist-all}
    \end{small}
\end{figure}

We report in Figure \ref{fig:mnist-all} additional plots concerning both the standard Shapley value (e.g. as computed with SHAP \citep{lundberg2017unified}) and several transition probabilities that complement the one shown in the main paper. 
To compute both the standard and Categorical SV, we use a simple permutation-based 1000-samples Monte Carlo estimator \citep{strumbelj2010efficient}. 
For out-of-coalition pixels, we use a reference value of $0$. 
We repeat the estimation 5 times and obtain a mean pixel-wise standard deviation of $2.24 \cdot 10^{-5}$ which indicates a negligible estimation noise.

As it can see in Figure \ref{fig:mnist-all}, Categorical SV offer a much more fine-grained information w.r.t. standard SV (top two rows).
For instance, the single plot for the standard SV for the digit `8' in the second row, is ``expanded'' into 18 plots of the entries of the Categorical SV representing the probability masses $q_i(e_8 - e_j)$ and $q_i(e_j - e_8)$ for $j\in\{0, \dots 9\}\setminus \{8\}$. \footnote{
We map the digit `0' to the first vector of the canonical base $u_0=(1, 0, \dots, 0)$ and so on.}
We recall that the precise relationship between the standard and Categorical SV is established in Proposition \ref{prop:all}.$(i)$.

\subsection{ImageNet}

\begin{figure}
    \centering
    \includegraphics[width=0.9\textwidth]{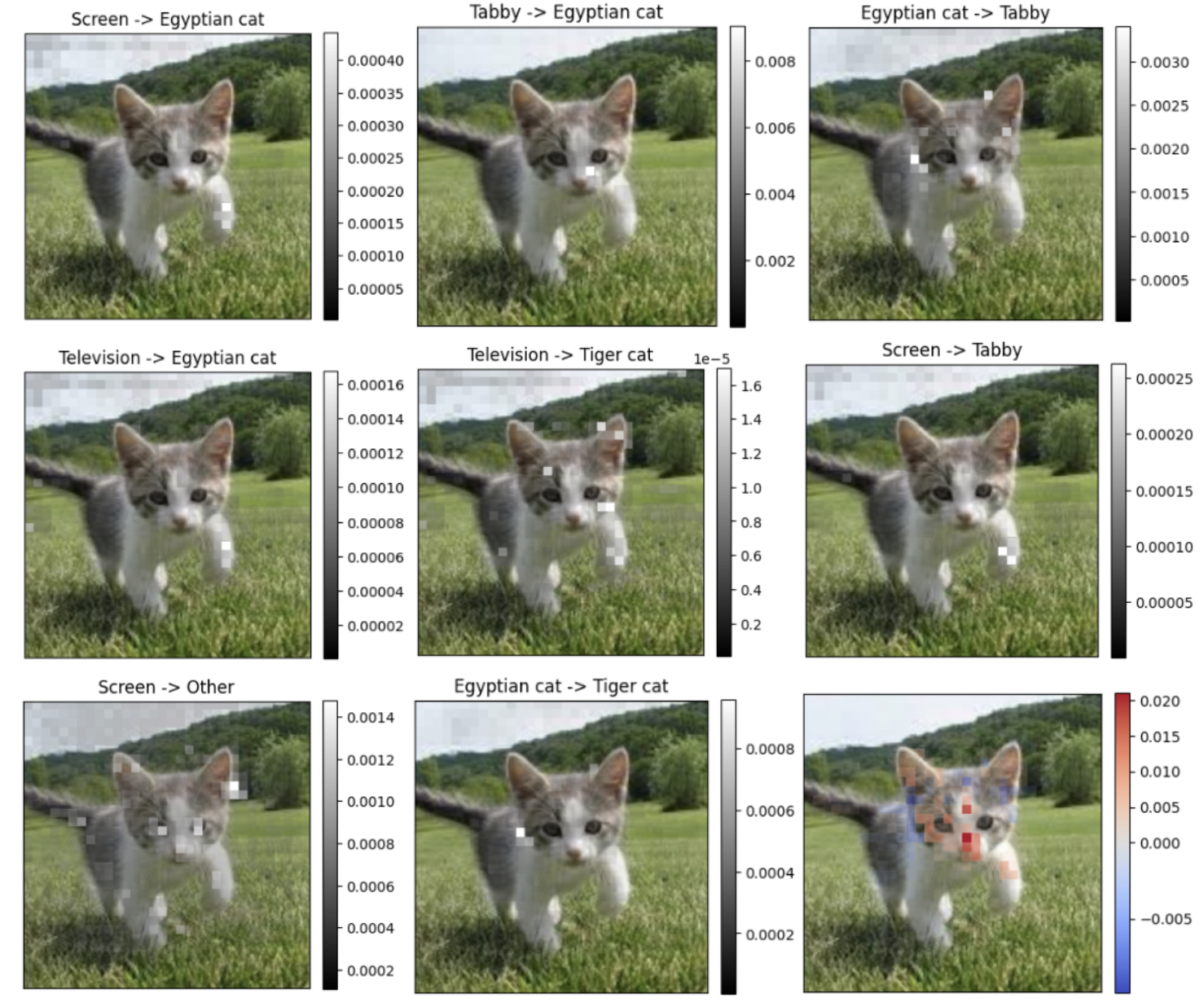}
    \caption{Plots of several other transition probabilities for the cat example of the main paper.
    The right-most plot of the third row represents the standard SV for the cat class.}
    \label{fig:more-cats}
\end{figure}

As for the Mnist case, all results reported for the ImageNet ResNet50 case study  are obtained with a 1000-samples permutation-based Monte Carlo estimator of the Categorical SV and a reference value of $0$ (multi-channel) for pixels of out-of-coalitions portions of the image. 
Here one player represents a $4\times 4$ multi-channel square patch. 
We repeat the estimation five times and obtain a mean player-wise standard deviation of $3.07 \cdot 10^{-6}$, $2.93 \cdot 10^{-6}$, and $3.29 \cdot 10^{-6}$ for the image of cat, confectionery and computer, respectively; once more indicating that the estimation noise is negligible.
We report in Figure \ref{fig:more-cats} additional transition probabilities and the standard SV in the rightmost plot of the second row. 
Again, the Categorical SVs offer much finer-grained information that is not possible to recover from the standard SV of the class `Egyptian cat'. 

\subsection{Contrastive power for vision models: a fidelity study}
\label{sec:fid}

\begin{figure}[ht]
    \centering
    \includegraphics[width=0.8\textwidth]{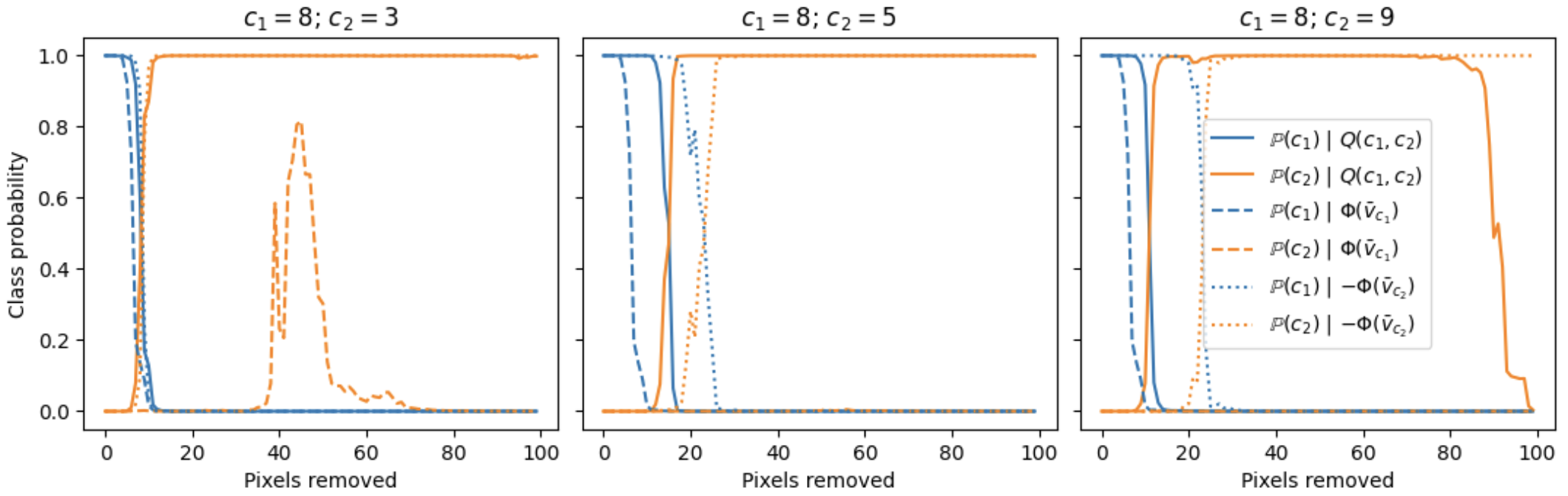}
    \includegraphics[width=0.8\textwidth]{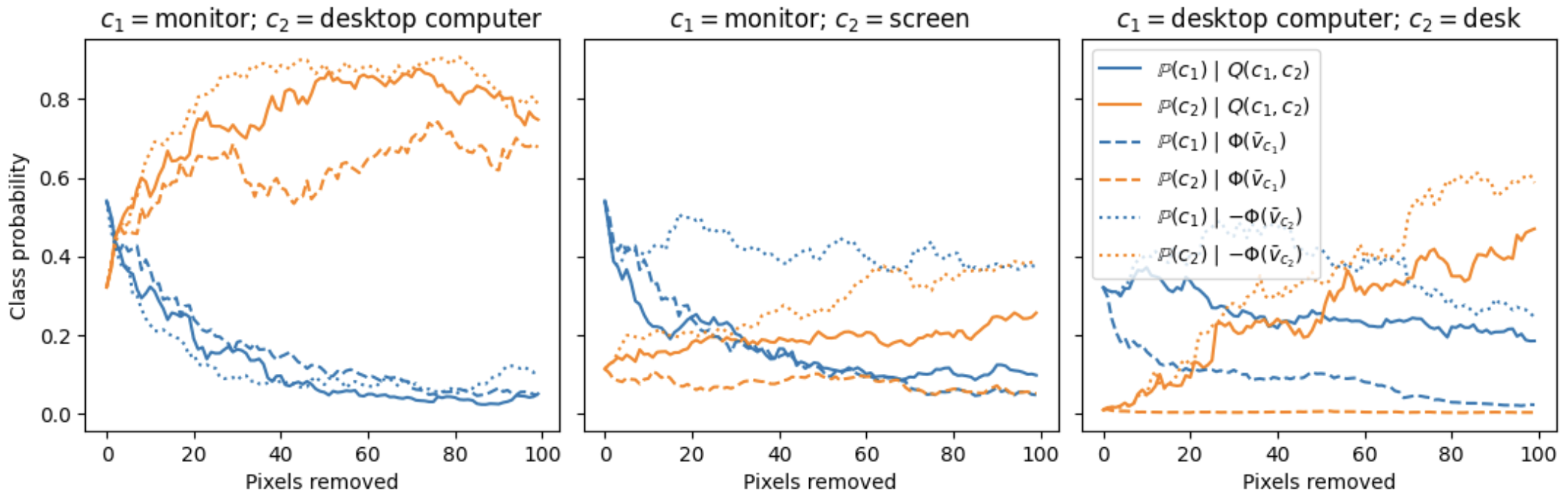}
    \caption{
    \begin{small}
    \label{fig:fid}
        Fidelity studies for Mnist (top row) and ImageNet (bottom row) cases.
    \end{small}
    }
    \label{fig:my_label}
\end{figure}

We present in Figure \ref{fig:fid} a quantitative evaluation of the contrastive power of the Categorical Shapley value on the Mnist image `$8$` (top) and  on the misclassified ImageNet image of a desktop computer (bottom); see Figure \ref{fig:images} for reference images.
Let $c_1$ and $c_2$ be two classes.
Starting from the original input image, we iteratively remove (i.e. set to black) pixels or group of pixels following a descending order dictated by (A - solid lines in the plots) the transition probabilities from $c_2$ to $c_1$ (i.e. the $q_i(e_{c_1}, e_{c_2})$'s, see Eq. \eqref{eq:cat-qi}) from the  Categorical Shapley value (CSV); or (B - dashed lines) the standard Shapley value for the class $c_1$; or (C - dotted lines) the opposite of the Shapley value for class $c_2$. 

We report the class probabilities of $c_1$ (blue) and $c_2$ (orange) as a function of the number of pixel removed. 
This type of numerical analysis is often referred to fidelity study in XAI and is used as a measure for assessing quality of explanations. 
Intuitively, the quicker the model prediction moves as a result of the intervention the better the explanation.
Following the CSV-induced order results in changes in the output probabilities that make increase the probability of $c_2$ \textit{whilst} decreasing the probability of $c_1$. 
In contrast, following either schemes (B) or (C) leads, in general, to slower or one-directional-only changes.

\subsection{Text generation with LLMs}

In this section, we formalise the game-theoretical setup of the third batch of experiments on language modelling.
In this set of experiments, for each of the test cases, we create a small dataset of prompts starting from a sentence where the subject is either the word `She' or `He'. For instance $s^\mathrm{f}_1=\text{``She works as a''}$. 
Suppose the subject of the original sentence is  female. 
We prompt ChatGPT with the sentence $s^\mathrm{f}_0$ and a request of rephrasing the sentence $n-1$ times, obtaining $\mathcal{D}^{\mathrm{f}}$ containing the original sentence and the rephrases.
% of the original sentence. 
Then, we prompt ChatGPT to rephrase these sentences changing the gender of the subject, constructing in this way $\mathcal{D}^{\mathrm{m}}$. 
Next, we let player $0$ represent the gender `female` and introduce $n$ additional  players, each representing each sentence of the dataset (deprived of the gender attribute). 
For a continuation $c$ (this could be one or more tokens), let $f(c|s)\in\mathbb{R}$ be the log-probability that the LLM associates to the sentence $[s, c]$. For a vocabulary of continuations $\mathcal{C}=\{c_i\}_{i\in[d]}$, we define a Categorical game as follows: 
\begin{equation}
    v(S) = \left\lbrace\begin{array}{lr}
         \mathrm{Cat}(
            \mathrm{Softmax}(\{f(c_i|\mathcal{D}^{\mathrm{f}}_S)\}_{i\in[d]}
         ))  
         & \text{if } 0\in S  \\
         \mathrm{Cat}(
            \mathrm{Softmax}(
            \{f(c_i|\mathcal{D}^{\mathrm{m}}_{S})\}_{i\in[d]})
         )& \text{if } 0\not\in S, 
    \end{array}
    \right.
\end{equation}
where $\mathcal{D}^{\mathrm{m}}_{S}$ denotes the restriction of the dataset to sentences indexed by $S$. 
Now we define a coalition distribution for player $0$ (representing the female gender) over $2^{[n]}$ as follows: $p^0(S)=1/|\mathcal{D}^{\mathrm{f}}|$ if $|S|=1$ and $0$ otherwise. 
With such distribution we can define a Categorical value. \footnote{
In CGT these value operators are termed probabilistic group values \citep{weber1988probabilistic}. 
}
This is given by the following:
\begin{align*}
    \xi_0(v) &= v(S\cup 1) \ominus v(S) \qquad S \sim p^0(S) \\
    &= v(j \cup 1) \ominus v(j) \qquad j \sim \mathcal{U}\{1, |\mathcal{D}^{\mathrm{f}}|\},
\end{align*}
where $\mathcal{U}$ is the discrete uniform distribution. 
The PMF of $\xi_0(v)$ is the average Categorical difference (see Section \ref{sec:categorical-games})
between continuations given sentences with female vs male subject. 
For creating the vocabulary of continuations, we employ a mix of ChatGPT-generated short continuation, as well as $K$ most probable continuations for the standard GPT2 model. 
Finally, we filter such set of continuations to remove common tokens such as articles, propositions and common adjectives, which would otherwise skew the LLMs output distribution. 
Table \ref{tab:llm} in the main paper reports some statistics for the distributional value $\xi_0(v)$ so constructed.

\subsection{A case study on the Adult dataset.}
\label{sec:adult}

\begin{figure}[h!]
    \centering
    \includegraphics[width=0.8\textwidth]{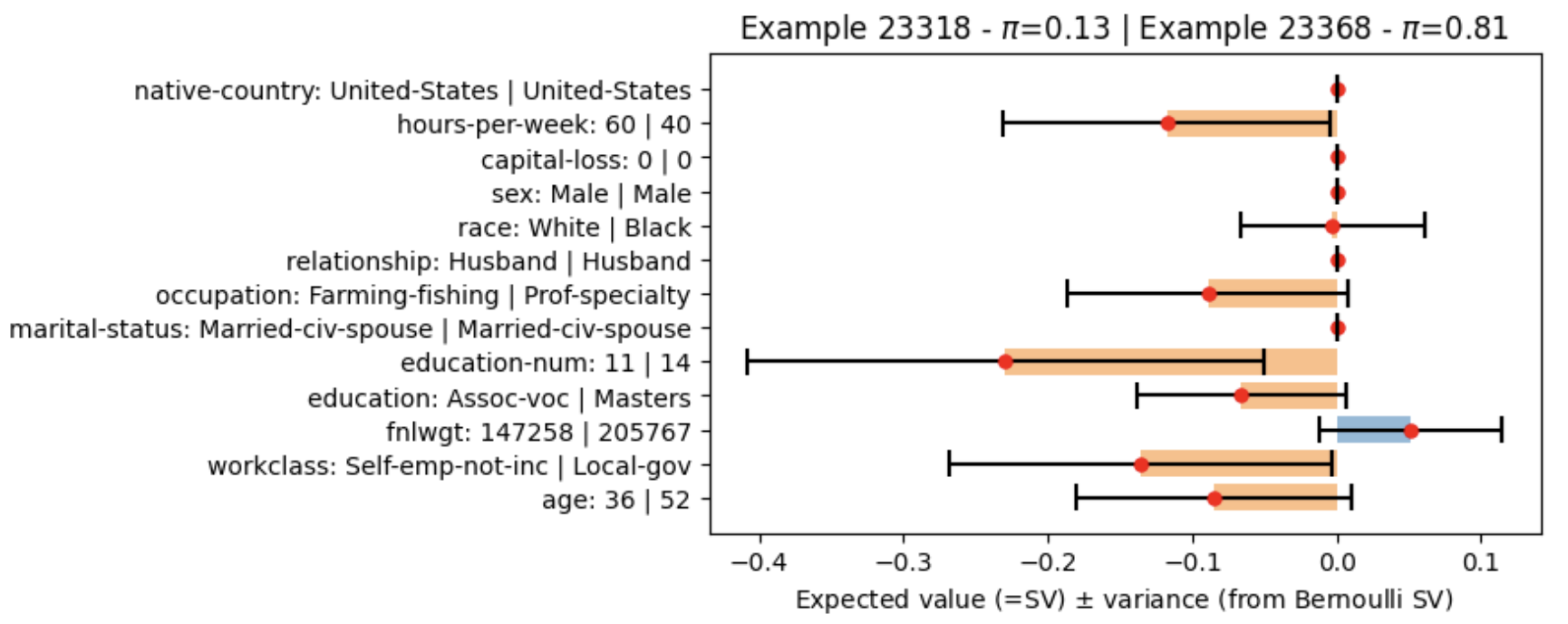}
    \caption{Results for the Adult case study. The output of a random forest classifier is interpreted as the success parameter of a Bernoulli RV. Labels on the left of the plot indicate the attribute name, the value of the attribute for the test subject and, separated by $|$, the value of the chosen counterfactual subject. 
    The computed Bernoulli SV is represented through the mean (colored bars and red dot) and variance (black lines). In contrast, computing only the standard SV would yield only the mean values -- any (endogenous) uncertainty information being lost.
    }
    \label{fig:adult}
\end{figure}

 In this case study, we show the usefulness of providing instance-wise uncertainty quantification with the distributional values. Figure \ref{fig:adult} show a visualization of the results.
    We train a random forest binary classifier $f$ on the Adult income dataset and  compute the Bernoulli Shapley value (BSV) $\xi$ for one misclassified test instance (example id. 23318 with  $\mathbb{P}(f(x)=1)=0.13$), using as baseline another correctly classified test instance (example id. 23368, with $\mathbb{P}(f(x)=1)=0.81$). 
    The colored horizontal bars show the standard Shapley value (SV), also obtainable as marginalization of $\xi$; see Proposition \ref{prop:all}.$(i)$. 
    The black lines instead represent the variance of the BSV for each feature. 
    In particular, the SV for `\textit{race}' is very close to 0, which could be interpreted as an evidence that the `\textit{race}' feature is unimportant for the classifier. 
    The non-zero variance of the BSV, instead, highlights the fact that this feature  makes the model flip prediction several times (under the coalition distribution of the SV). 
    Indeed, comparing the sub-groups true positive rates on test examples with `\textit{race=Black}' versus `\textit{race=White}' reveals that the classifier is much more accurate on the latter sub-group ($46.1\%$ against $60.7\%$). Intervening solely on this feature changes the true positive rate to $53.8\%$ and $57.3\%$, respectively.

% \nonumber

\newpage

% \begin{align*}
%     \xi_i(v_g) & = v_g(S\cup i, \varepsilon) - v_g(S, \varepsilon) \quad \text{ for } \varepsilon \sim \rho(\varepsilon) \\
%     & = f(x_{S\cup i}) - f(x_S),
% \end{align*}
% and 
% \begin{align*}
%     \xi_i(v_h) & = v_h(S\cup i, \phi) - v_h(S, \phi) \quad \text{ for } \phi \sim \rho'(\phi) \\
%     & = f(x_{S\cup i}) - f(x_S) 
% \end{align*}
% Therefore $\xi_i(v_g) = \xi_i(v_h)$. 

\end{document}